\documentclass{article}
\usepackage{a4wide}
\usepackage{amsthm}
\usepackage{xcolor}
\usepackage{setspace}
\usepackage{amssymb}
\usepackage[T1]{fontenc}

\usepackage{amsmath,amsfonts,bm}

\def\eqref#1{equation~\ref{#1}}

\def\1{\bm{1}}

\DeclareMathAlphabet{\mathsfit}{\encodingdefault}{\sfdefault}{m}{sl}
\SetMathAlphabet{\mathsfit}{bold}{\encodingdefault}{\sfdefault}{bx}{n}

\newcommand{\E}{\mathbb{E}}

\DeclareMathOperator*{\argmax}{arg\,max}
\DeclareMathOperator*{\argmin}{arg\,min}

\usepackage[pagebackref]{hyperref}
  \definecolor{mydarkblue}{rgb}{0,0.08,0.45}
\hypersetup{colorlinks,linkcolor={mydarkblue},citecolor={mydarkblue},urlcolor={red}}  
\usepackage{url}

\usepackage{thmtools}
\usepackage{thm-restate}

\usepackage{algorithm}
\usepackage{algpseudocode}
\usepackage{adjustbox}
\usepackage{tensor}
\usepackage{booktabs}
\usepackage{dsfont}

\usepackage{natbib}
\title{An Investigation of Offline Reinforcement Learning in \\ Factorisable Action Spaces}

\author{
    Alex Beeson$^\mathbf{1,2}$, David Ireland$^\mathbf{1}$, Giovanni Montana$^\mathbf{1,3,4}$ \\
    $^1$Warwick Manufacturing Group, University of Warwick, Coventry, UK \\
    $^2$Warwick Medical School, University of Warwick, Coventry, UK \\
    $^3$Department of Statistics, University of Warwick, Coventry, UK \\
    $^4$Alan Turing Institute, London, UK \\
    \texttt{\{alex.beeson, david.ireland, g.montana\}@warwick.ac.uk}
}

\begin{document}

\maketitle

\begin{abstract}
Expanding reinforcement learning (RL) to offline domains generates promising prospects, particularly in sectors where data collection poses substantial challenges or risks.  Pivotal to the success of transferring RL offline is mitigating overestimation bias in value estimates for state-action pairs absent from data. Whilst numerous approaches have been proposed in recent years, these tend to focus primarily on continuous or small-scale discrete action spaces.  Factorised discrete action spaces, on the other hand, have received relatively little attention, despite many real-world problems naturally having factorisable actions.  In this work, we undertake a formative investigation into offline reinforcement learning in factorisable action spaces.  Using value-decomposition as formulated in DecQN as a foundation, we present the case for a factorised approach and conduct an extensive empirical evaluation of several offline techniques adapted to the factorised setting.  In the absence of established benchmarks, we introduce a suite of our own comprising datasets of varying quality and task complexity.  Advocating for reproducible research and innovation, we make all datasets available for public use alongside our code base.
\end{abstract}

\section{Introduction}
The idea of transferring the successes of reinforcement learning (RL) to the offline setting is an enticing one. The opportunity for agents to learn optimal behaviour from sub-optimal data prior to environment interaction extends RL's applicability to domains where data collection is costly, time-consuming or dangerous \citep{lange2012batch}. This includes not only those domains where RL has traditionally found favour, such as games and robotics \citep{mnih2013playing, hessel2018rainbow, kalashnikov2018scalable, mahmood2018benchmarking}, but also areas in which online learning presents significant practical and/or ethical challenges, such as autonomous driving \citep{RLAutoDriving} and healthcare \citep{yu2021reinforcement}. 

Unfortunately, taking RL offline is not as simple as naively applying standard off-policy algorithms to pre-existing datasets.  A substantial challenge arises from the compounding and propagation of overestimation bias in value estimates for actions absent from data \citep{fujimoto2019off}. This bias stems from the underlying bootstrapping procedure used to derive such estimates and subsequent maximisation to obtain policies, whether implicit such as in Q-learning or explicit as per actor-critic methods \citep{levine2020offline}.  Fundamentally, agents find it difficult to accurately determine the value of actions not previously encountered, and thus any attempt to determine optimal behaviour based on these values is destined to fail.  Online, such inaccuracies can be compensated for through continual assimilation of environmental feedback, but offline such a corrective mechanism is no longer available.

In response to these challenges, there has been a plethora of approaches put forward that aim to both curb the detrimental effects of overestimation bias as well as let agents discover policies that improve over those that collected the data to begin with \citep{levine2020offline}. The last few years in particular have seen a wide variety of approaches proposed, making use of policy constraints \citep{fujimoto2019off, zhou2021plas, wu2019behavior, kumar2019stabilizing, kostrikov2021offlineImp, fujimoto2021minimalist}, conservative value estimation \citep{kostrikov2021offline, kumar2020conservative}, uncertainty estimation \citep{an2021uncertainty, ghasemipour2022so, baipessimistic, yangrorl, nikulin2023anti, beeson2024balancing} and environment modelling \citep{kidambi2020morel, yu2021combo, argenson2020model, janner2022planning, yu2020mopo, swazinna2021overcoming}, to name just a few. Each approach comes with its own strengths and weaknesses in terms of performance, computational efficiency, ease of implementation and hyperparameter tuning. 

To date, most published research in offline RL has focused on either continuous or small-scale discrete action spaces. However, many complex real-world problems can be naturally expressed in terms of \emph{factorised action spaces}, where global actions consist of multiple distinct sub-actions, each representing a key aspect of the decision process. Examples include ride-sharing \citep{lin2018efficient}, recommender systems \citep{zhao2018deep}, robotic assembly \citep{driess2020deep} and healthcare \citep{liu2020reinforcement}. 

Formally, the factorised action space is considered the Cartesian product of a finite number of independent discrete action sets, i.e. $\mathcal{A} = \mathcal{A}_1 \times ... \times \mathcal{A}_N$, where $\mathcal{A}_i$ contains $n_i$ (sub-)actions and $N$ corresponds to the dimensionality of the action space. The total number of actions, often referred to as atomic actions, is thus $\prod_{i=1}^N n_i$, which can undergo combinatorial explosion if $N$ and/or $n_i$ grow large.  For example, in robotics a set of actions could correspond to moving a joint up, down or not at all, i.e. $n_i=3$.  For machines with multiple joints the global action is the set of individual actions for each joint.  The total number of possible global actions is thus $3^N$ which grows exponentially as the number of joints increases.

In recognition of this possibility, various strategies in the online setting have been devised to preserve the effectiveness of commonly used discrete RL algorithms \citep{tavakoli2018action, tang2020discretizing}.  The concept of \emph{value-decomposition} \citep{seyde2022solving} is particularly prominent, wherein value estimates for each sub-action space are computed independently, yet are trained to ensure that their aggregate mean converges towards a universal value.  The overall effect is to reduce the total number of actions for which a value needs to be learnt from a product to a sum, making problems with factorisable actions spaces much more tractable for approaches such as Q-learning.  Referring back to the robotics example, by using a decomposed approach the number of actions requiring value estimates is reduced from $3^N$ to $3N$. 

In this work, we undertake an initial investigation into offline RL in factorisable action spaces.  Using value-decomposition, we show how a factorised approach provides several benefits over a standard atomic action representation.  We conduct an extensive empirical evaluation of a number of offline approaches adapted to the factorised setting using a newly created benchmark designed to test an agent's ability to learn complex behaviours from data of varying quality.  In the spirit of advancing research in this area, we provide open access to these datasets as well as our full code base: \url{https://github.com/AlexBeesonWarwick/OfflineRLFactorisableActionSpaces}.

To the best of our knowledge, this investigation represents the first formative analysis of offline RL in factorisable action spaces.  We believe our work helps pave the way for developments in this important domain, whilst also contributing to the growing field of offline RL more generally.

\section{Preliminaries}\label{Prelim}

\subsection{Offline reinforcement learning}
    Following standard convention, we begin by defining a Markov Decision Process (MDP) with state space $\mathcal{S}$, action space $\mathcal{A}$, environment dynamics $T(s' \mid s, a)$, reward function $R(s, a)$ and discount factor $\gamma \in [0, 1]$ \citep{sutton2018reinforcement}.  An agent interacts with this MDP by following a state-dependent policy $\pi (a \mid s)$, with the primary objective of discovering an optimal policy $\pi^* (a \mid s)$ that maximises the expected discounted sum of rewards, $\E_\pi\left[\sum_{t=0}^\infty\gamma^t r(s_t, a_t)\right]$.

    A popular approach for achieving this objective is through the use of Q-functions, $Q^\pi (s, a)$, which estimate the value of taking action $a$ in state $s$ and following policy $\pi$ thereafter.  In discrete action spaces, optimal Q-values can be obtained by repeated application of the Bellman optimality equation:
    $$
    Q^*(s, a) = r(s, a) + \gamma \E_{s' \sim T} \left[\max_{a'} Q^*(s', a')\right] \; .
    $$
    These Q-values can then be used to define an implicit policy such that $\pi(s) = \argmax_a Q(s,a)$ i.e. the action that maximises the optimal Q-value at each state.  
    
    Given the scale and complexity of real-world tasks, Q-functions are often parameterised \citep{mnih2013playing}, with learnable parameters $\theta$ that are updated so as to minimise the following loss:    
    \begin{equation}\label{BellmanOptimality}
    L(\theta) = \frac{1}{|B|} \sum_{(s, a, r, s') \sim \mathcal{B}} \left(Q_\theta(s, a) - y(r, s')\right)^2 \; ,
    \end{equation}
    where $y(r, s') = r + \gamma \max_{a'} Q_\theta(s', a')$ is referred to as the target value, and $B$ denotes a batch of transitions sampled uniformly at random from a replay buffer $\mathcal{B}$.  To promote stability during training, when calculating Q-values in the target it is common to use a separate target network $Q_{\hat{\theta}}(s', a')$ \citep{mnih2013playing, hessel2018rainbow} with parameters $\hat{\theta}$ updated towards $\theta$ either via a hard reset every specified number of steps, or gradually using Polyak-averaging.
  
    In the offline setting, an agent is prohibited from interacting with the environment and must instead learn solely from a pre-existing dataset of interactions $\mathcal{B}=(s_b, a_b, r_b, s'_b)$, collected from some unknown behaviour policy (or policies) $\pi_\beta$ \citep{lange2012batch}.  This lack of interaction allows errors in Q-value estimates to compound and propagate during training, often resulting in a complete collapse of the learning process \citep{fujimoto2019off}.  Specifically, Q-values for out-of-distribution actions (i.e. those absent from the  dataset) suffer from overestimation bias as a result of the maximisation carried out when determining target values \citep{thrun1993issues, gaskett2002q}.  The outcome is specious Q-values estimates, and policies derived from those estimates consequently being highly sub-optimal.  In order to compensate for this overestimation bias, Q-values must be regularised by staying ``close'' to the source data \citep{levine2020offline}.  
    
    \subsection{Decoupled Q-Network}\label{Prelim-DecQN}  
    By default, standard Q-learning approaches are based on atomic representations of action spaces \citep{sutton2018reinforcement}.  This means that, in a factorisable action space, Q-values must be determined for every possible combination of sub-actions. This potentially renders such approaches highly ineffective due to combinatorial explosion in the number of atomic actions. Recalling that the action space can be thought of as a Cartesian product, then for each $\mathcal{A}_i$ we have that $|\mathcal{A}_i| = n_i$, and so the total number of atomic actions is $\prod^N_{i=1}n_i$. This quickly grows unwieldly as the number of sub-action spaces $N$ and/or number of actions within each sub-action space $n_i$ increase.

    To address this issue, the Branching Dueling Q-Network (BDQ) proposed by \citet{tavakoli2018action} learns value estimates for each sub-action space independently and can be viewed as a single-agent analogue to independent Q-learning from multi-agent reinforcement learning (MARL) \citep{claus1998dynamics}. \citet{seyde2022solving} expand on this work with the introduction of the Decoupled Q-Network (DecQN), which computes value estimates in each sub-action space independently, but learns said estimates such that their mean estimates the Q-value for the combined (or  global) action. Such an approach is highly reminiscent of the notion of value-decomposition used in cooperative MARL \citep{sunehag2017value, rashid2020monotonic, rashid2020weighted, du2022value}, with sub-action spaces resembling individual agents.
    
    In terms of specifics, DecQN introduces a utility function $U^i_{\theta_i}(s, a_i)$ for each sub-action space and redefines the Q-value to be:
    \begin{equation}\label{Q-Decomp}
        Q_\theta(s, \mathbf{a}) = \frac{1}{N} \sum^N_{i=1} U^i_{\theta_i}(s, a_i) \; , 
    \end{equation}
    where $\mathbf{a} = (a_1, ..., a_N)$ is the global action, $\theta_i$ are the parameters for the $i$th utility function and $\theta = \{\theta_i\}_{i=1}^N$ are the global set of parameters. The loss in Equation (\ref{BellmanOptimality}) is updated to incorporate this utility function structure:
    \begin{equation}\label{DecQN}
    L(\theta) = \frac{1}{|B|} \sum_{(s, \mathbf{a}, r, s') \sim \mathcal{B}} \left(Q_\theta(s, \mathbf{a}) - y(r, s')\right)^2 \; , 
    \end{equation}
    where 
    $$
    y(r, s') = r + \frac{\gamma}{N} \sum^N_{i=1} \max_{a'_i} U^i_{\theta_i}(s', a'_i) \; .
    $$

    As each utility function only needs to learn about actions within its own sub-action space, this reduces the total number of actions for which a value must be learnt to $\sum^N_{i=1}n_i$, thus preserving the functionality of established Q-learning algorithms.     Whilst there are other valid decomposition methods, in this work we focus primarily on the decomposition proposed in DecQN. In Appendix \ref{sec: decomposition comparisons} we provide a small ablation justifying our choice. 

\section{Related Work}\label{RelatedWork}

    \subsection{Offline RL}      
        Numerous approaches have been proposed to help mitigate Q-value overestimation bias in offline RL.  In BCQ \citep{fujimoto2019off}, this is achieved by cloning a behaviour policy and using generated actions to form the basis of a policy which is then optimally perturbed by a separate network.  BEAR \citep{kumar2019stabilizing}, BRAC \citep{wu2019behavior} and Fisher-BRC \citep{kostrikov2021offline} also make use of cloned behaviour policies, but instead use them to minimise divergence metrics between the learned and cloned policy.  One-step RL \citep{brandfonbrener2021offline, gulcehre2020addressing} explores the idea of combining fitted Q-evaluation and various policy improvement methods to learn policies without having to query actions outside the data.  This is expanded upon in Implicit Q-learning (IQL) \citep{kostrikov2021offlineImp}, which substitutes fitted Q-evaluation with expectile regression.  TD3-BC \citep{fujimoto2021minimalist} adapts TD3 \citep{fujimoto2018addressing} to the offline setting by directly incorporating behavioural cloning into policy updates, with TD3-BC-N/SAC-BC-N \citep{beeson2024balancing} employing ensembles of Q-functions for uncertainty estimation to alleviate issues relating to overly restrictive constraints as well as computational burden present in other ensembles based approaches such as SAC-N \& EDAC \citep{an2021uncertainty}, MSG \citep{ghasemipour2022so}, PBRL \citep{baipessimistic} and RORL \citep{yangrorl}.  

        In the majority of cases the focus is on continuous action spaces, and whilst there have been adaptations and implementations in discrete action spaces \citep{fujimoto2019benchmarking, gu2022learning}, these tend to only consider a small number of (atomic) actions.  This is also reflected in benchmark datasets such as D4RL \citep{fu2020d4rl} and RL Unplugged \citep{gulcehre2020rl}.  Our focus is on the relatively unexplored area of factorisable discrete action spaces.  
    
    \subsection{Action decomposition}

        Reinforcement learning algorithms have been extensively studied in scenarios involving large, discrete action spaces.  In order to overcome the challenges inherent in such scenarios, numerous approaches have been put forward based on action sub-sampling \citep{van2020q, hubert2021learning}, action embedding \citep{dulac2015deep, gu2022learning} and curriculum learning \citep{farquhar2020growing}.  However, such approaches are tailored to handle action spaces comprising numerous atomic actions, and do not inherently tackle the complexities nor utilise the structure posed by factorisable actions.
        
        For factorisable action spaces various methods have been proposed, such as learning about sub-actions independently via value-based \citep{sharma2017learning, tavakoli2018action} or policy gradient methods \citep{tang2020discretizing, seyde2021bang}. Others have also framed the problem of action selection in factorisable action spaces as a sequence prediction problem, where the sequence consists of the individual sub-actions \citep{metz2017discrete, pierrotfactored, chebotar2023q}.  
        
        There exists a strong connection between factorisable action spaces and MARL, where the selection of a sub-action can be thought of as an individual agent choosing its action in a multi-agent setting. Value-decomposition has been shown to be an effective approach in MARL \citep{sunehag2017value, rashid2020monotonic, rashid2020weighted, du2022value}, utilising the \textit{centralised training with decentralised execution} paradigm \citep{kraemer2016multi}, which allows agents to act independently but learn collectively. DecQN \citep{seyde2022solving} and REValueD \citep{ireland2023revalued} have subsequently shown such ideas can be used with factorised action spaces in single-agent reinforcement learning, demonstrating strong performance on a range of tasks that vary in complexity.  Theoretical analysis of errors in Q-value estimates has also been conducted in efforts to stabilise training \citep{ireland2023revalued, thrun1993issues}.
        
        In this work, we focus on adapting DecQN to the offline setting by incorporating existing offline techniques. Whilst prior work has explored offline RL with value decomposition \citep{tang2022leveraging}, this was limited to specific low-dimensional healthcare applications using only BCQ. Furthermore, accurately evaluating performance in such domains is notorious challenging \citep{gottesman2018evaluating}. In contrast, we systematically study multiple offline methods using low and high-dimensional factorised action spaces across a suite of benchmark tasks.

\section{A case for factorisation and decomposition in offline RL}
    \label{Case}
    As mentioned in Section \ref{Prelim-DecQN}, value-decomposition provides a mechanism for overcoming challenges in standard Q-learning arising from exponential growth in the number of atomic actions.  This can be seen most clearly from a practical standpoint, in which the number of actions that require value estimation is significantly reduced when moving from atomic to factorised action representation (see Table \ref{tab:dm_control_suite} for example).  However, even in cases where atomic representation remains feasible, there are still several benefits to a factorised and decomposed approach which are particularly salient in the offline case.  

    To see this, we begin by making the assumption that Q-value estimates from function approximation $Q_\theta(s, \mathbf{a})$ carry some noise $\epsilon(s, \mathbf{a})$ \citep{thrun1993issues}.  This noise can be modelled as a random variable, such that $Q_\theta(s, \mathbf{a}) = Q^\pi(s, \mathbf{a}) + \epsilon(s, \mathbf{a})$, where $Q^\pi(s, \mathbf{a})$ is the true but unknown Q-value for policy $\pi$.  For clarity, we emphasise this noise stems from the fact the function approximator cannot represent the true Q-function precisely, as opposed to other factors such as sampling variation.  As noted by \citet{thrun1993issues}, the expected overestimation will be maximal if all actions for a particular state share the same target Q-value.  Incorporating all these ideas into the Q-learning framework we can define the target difference as
    \begin{equation}\label{DQN_single}
        \begin{split}
            Z_{s'}^{dqn} & = \gamma \Big(\max_{\mathbf{a}'} Q_\theta(s', \mathbf{a}') - \max_{\mathbf{a}'} Q^\pi(s', \mathbf{a}')  \Big); \\
                         & = \gamma \Big(\max_{\mathbf{a}'} \epsilon(s', \mathbf{a'})  \Big).
        \end{split}
    \end{equation}

    To incorporate decomposition as defined in Equation \ref{Q-Decomp}, the above can be extended to utility value estimates \citep{ireland2023revalued}.  Now we assume that utility value estimates from function approximation $U_{\theta_i}^i (s, a_i)$ carry some noise $\epsilon^i(s, a_i)$, such that $U_{\theta_i}^i (s, a_i) = U_i^{\pi_i} (s, a) + \epsilon^i(s, a_i)$, where $U_i^{\pi_i} (s, a_i)$ is the true but unknown utility value for policy $\pi_i$.  Under the additional assumption that true Q-values decompose in the same way as Equation \ref{Q-Decomp} \citep{ireland2023revalued}, we can define the target difference under DecQN as
    \begin{equation}\label{DecQN_single}
        \begin{split}
            Z_{s'}^{dec} & = \gamma \biggl( \frac{1}{N} \sum_{i=1}^N \max_{a'_i} U_{\theta_i}^i(s', a'_i) - \frac{1}{N} \sum_{i=1}^N \max_{a'_i} U_i^{\pi_i}(s', a'_i) \biggl); \\
                         & = \gamma \biggl( \frac{1}{N} \sum_{i=1}^N \max_{a'_i} \epsilon^i(s', a'_i) \biggl).
        \end{split}
    \end{equation}

\begin{figure}[ht]
            \centering
            \includegraphics[width=12.5cm]{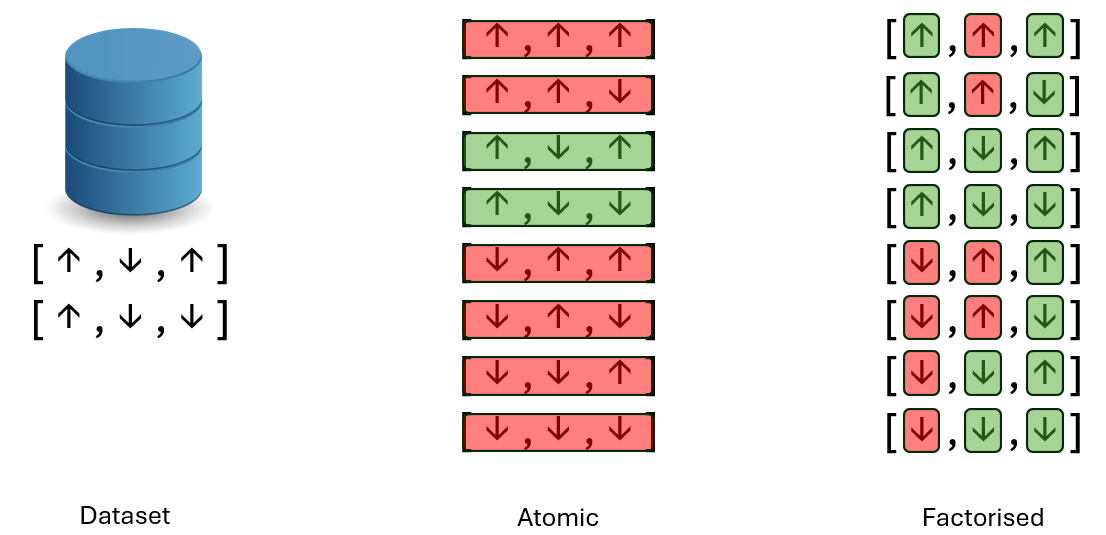}
            \caption{In this simple example there are $N=3$ sub-action dimensions, each with two actions \{$\uparrow$, $\downarrow$\}.  In-distribution and out-of-distribution actions/sub-actions are highlighted in green and red, respectively.  For a particular state, the dataset contains two global actions.  Under atomic representation only actions which match those in the dataset are in-distribution.  Under factorised representation, individual sub-actions which match those in the dataset are in-distribution.  Atomic actions that are out-of-distribution can contain sub-actions that are in-distribution when factorised.}
            \label{fig: AtomVsFactEg}
        \end{figure}

    We can expand on each of these by considering the composition of global actions in terms of in-distribution and out-of-distribution sub-actions.  When actions are represented atomically, a particular global action $\mathbf{a}$ is either in-distribution or out-of-distribution, depending on its presence (or absence) in the dataset.  However, under a factorised representation a global action $\mathbf{a}$ is composed of sub-actions $(a_1, a_2, ..., a_N)$, and hence it is the individual sub-actions that are either in-distribution or out-of-distribution.  This means that global actions which are out-of-distribution can contain individual sub-actions that are in-distribution when factorised, as illustrated in Figure \ref{fig: AtomVsFactEg}.

    In recognition of this property, we now differentiate between errors in value estimates for in-distribution and out-of-distribution actions.  Using the precept that out-of-distribution errors are inherently larger than in-distribution errors, let $\epsilon^{in}(s, \mathbf{a}^{in})$, $\epsilon^{in}(s, a^{in}_i)$ and  $\epsilon^{out}(s, \mathbf{a}^{out})$, $\epsilon^{out}(s, a^{out}_i)$ denote the errors in Q-value/utility value estimates for in-distribution $\mathbf{a}^{in}, a^{in}_i$ and out-of-distribution $\mathbf{a}^{out}, a^{out}_i$ global/sub-actions, respectively.  The target difference for DQN and DecQN under this framework become
    \begin{align}
        Z_{s}^{dqn} &= \gamma \left(\max \{ \max_{\mathbf{a}^{in}} \epsilon^{in}(s, \mathbf{a}^{in}), \max_{\mathbf{a}^{out}} \epsilon^{out}(s, \mathbf{a}^{out})\} \right) \;; \\
        Z_{s}^{dec} &= \gamma \left( \frac{1}{N} \sum_{i=1}^N \max \{ \max_{a_i^{in}} \epsilon^{in}(s, a^{in}_i), \max_{a_i^{out}} \epsilon^{in}(s, a^{out}_i) \} \right) \; ;
    \end{align}
    where we have dropped the prime symbol to avoid overloading notation.  Now the maximum is taken over a pooled sample of errors from two different distributions as opposed to a single distribution in Equations \ref{DQN_single} and \ref{DecQN_single}.

    The important point to note here is that in both cases the target difference is now dependent on the relative number of in-distribution and out-of-distribution actions/sub-actions.  Since errors from in-distribution actions/sub-actions are smaller than for out-of-distribution, the larger the number of in-distribution actions/sub-actions the smaller the expected overestimation.  In the offline setting, we cannot change the status of an action/sub-action from out-of-distribution to in-distribution through environment interaction.  However, as noted above, we can improve the coverage of sub-actions through factorisation.  Hence, for the same dataset we can potentially reduce overestimation bias moving from an atomic action representation to a factorised one.  This is particularly beneficial in the offline setting where issues relating to overestimation bias primarily stem from Q-value estimates for out-of-distribution actions.  

    In Appendix \ref{sec: uniform_noise} we provide support for the ideas presented in this Section by empirically assessing the properties of target differences under DQN and DecQN for uniformly distributed noise. 

\subsection{Appropriateness of decomposition in the offline setting}\label{Case: Approp}
    The previous analysis assumes both the approximate and true Q-function decompose as per Equation \ref{Q-Decomp}.  In general, such a decomposition is unlikely to perfectly capture intricate aspects of an environment.  Nonetheless, such a decomposition has been shown to be effective in the online setting \citep{ireland2023revalued}, and we note that under the following conditions it can act as reasonable approximation, allowing us to retain the aforementioned benefits of improved action/sub-action coverage.  

    \subsubsection*{Weak inter-action dependence}
    If the effect of each sub-action on the reward or the next state is relatively independent of other sub-actions, the decomposition is more likely to hold. This condition aligns with the notion of action independence in factored MDPs, where the transition dynamics can be decomposed across sub-actions with minimal cross-interaction. In such cases, the global Q-function can be closely approximated by a sum of local utility functions. This can also be tied to work on factored MDPs \citep{kearns1999efficient}, where state and action space factorisations allow efficient policy computation under certain independence assumptions. Of course, true independence of sub-actions in an MDP might be rare in practice, especially when complex dependencies exist between sub-actions (e.g., coordination between multiple control signals).  Still, if dependencies are weak, factorisation can be a good approximation. It's also important to emphasise this only holds under certain independence assumptions that often arise in factored MDPs but may not be present in more general MDPs.

    \subsubsection*{Approximately factorisable reward structures}
    The decomposition is also reasonable when the reward function can be roughly expressed as a sum over individual sub-actions, which is again related to the literature on reward decomposition in factored MDPs \citep{guestrin2003efficient}. In scenarios where the global reward is a combination of local rewards for each sub-action, factorisation provides a natural approximation for value function decomposition. The caveat is that, in many real-world scenarios, rewards are not perfectly factorisable, and there could be interactions between sub-actions that affect the total reward in a non-linear way. Thus, such an approximation may break down in complex environments with highly entangled reward structures. However for certain structured problems, especially those with weak interactions, this approximation can still yield practical benefits.

    \subsection{Limitations of decomposition in the offline setting}\label{Case: Limit}
    It is important to highlight that while the benefits of a decomposed approach in the offline setting can offset shortcomings in modelling inter-action dependence, there are limitations.  For settings where sub-actions are strongly dependent, the value of a global action may differ significantly to the average value of its constituent utilities, introducing errors that are no longer outweighed by the reduction in overestimation bias from factorisation/decomposition.

    As an example, consider the phenomena of pharmacodynamic and pharmacokinetic drug interactions.  In a healthcare setting, it is common for patients to receive multiple drugs as part of a treatment regime, with dosages and routes adjusted based on how the patient responds.  If our goal is to train an offline RL to optimise treatment regimes, the level of interaction (positive or negative) between drugs/dosages/route would dictate whether a factorised/decomposed approach would be appropriate.  If there is little to no interaction then a factorised/decomposed approach could be effective.  If there are strong interactions a factorised/decomposed approach would likely fail to capture these important characteristics and hence be much less effective.

   
\section{Algorithms}\label{algos}
    In this Section we introduce several approaches for adapting DecQN to the offline setting based on existing techniques.  We focus on methods that offer distinct takes on combatting overestimation bias, namely, policy constraints, conservative value estimation, implicit Q-learning and one-step RL.  In each case, attention shifts from Q-values to utility values, with regularisation performed at the sub-action level.  The full procedure for each algorithm can be found in Appendix \ref{appendix:algos}

\subsection{DecQN-BCQ}
    Batch Constrained Q-learning (BCQ) \citep{fujimoto2019off, fujimoto2019benchmarking} is a policy constraint approach to offline RL.  To compensate for overestimation bias for out-of-distribution actions, a cloned behaviour policy $\pi_\phi$ is used to restrict the actions available for target Q-values estimates, such that their probability under the behaviour policy meets a relative threshold $\tau$.  This can be adapted and integrated into DecQN by cloning separate behaviour policies $\pi^i_{\phi_i}$ for each sub-action dimension and restricting respective sub-actions available for corresponding target utility value-estimates.  The target value from Equation (\ref{DecQN}) becomes:
    \[
    y(r, s') = r + \frac{\gamma}{N} \sum^N_{i=1} \max_{a'_i \; : \; \rho^i(a'_i) \geq \tau} U^i_{\theta_i}(s', a'_i) \; , 
    \]
    where 
    $
    \rho^i(a'_i) = \pi^i_{\phi_i} (a'_i \mid s') / \max_{\hat{a}'_i} \pi^i_{\phi_i} (\hat{a}'_i \mid s') \; 
    $
    is the relative probability of sub-action $a'_i$ under policy $\pi^i_{\phi_i}$.  Each cloned behaviour policy is trained via supervised learning with $\phi = \{\phi\}_{i=1}^N$.  The full procedure can be found in Algorithm \ref{alg:DecQN+BCQ}.

\subsection{DecQN-CQL}
    Conservative Q-learning (CQL) \citep{kumar2020conservative} attempts to combat overestimation bias by targeting Q-values directly.  The loss in Equation (\ref{BellmanOptimality}) is augmented with a term that ``pushes-up'' on Q-value estimates for actions present in the dataset and ``pushes-down'' for all others.  Under one particular variant this additional loss takes the form
    \[
    L_{CQL}(\theta) = \frac{\alpha}{|B|} \sum_{s, a \sim \mathcal{B}} \big[ \log \sum_{a_i \in A} \exp(Q_\theta(s,a_i)) - Q_\theta(s,a) \big] \; ;
    \]
    which equates to performing behavioural cloning using log-likelihood when the policy is a softmax over Q-values \citep{luo2023action} since
    \begin{equation*}
    \begin{split}
        \frac{\alpha}{|B|} \sum_{s, a \sim \mathcal{B}} \big[ \log \sum_{a_i \in A} \exp (Q_\theta(s,a_i)) - Q_\theta(s,a) \big] & = -\frac{\alpha}{|B|} \sum_{s, a \sim \mathcal{B}} \Big[ \log \frac{\exp(Q_\theta(s,a))}{\sum_{a_i \in A} \exp (Q_\theta(s,a_i))} \Big] \\
        & = -\frac{\alpha}{|B|} \sum_{s, a \sim \mathcal{B}}  \log \pi_\theta(a \mid s).
    \end{split}
    \end{equation*}
    
    This variant can be adapted and incorporated into DecQN by ``pushing-up'' on utility value estimates for sub-actions present in data and ``pushing-down'' for all others.  The additional loss under DecQN becomes:
    \begin{equation}\label{CQL_Loss}
    L_{CQL}(\theta) = \frac{\alpha}{|B|} \sum_{s, \mathbf{a} \sim \mathcal{B}} \frac{1}{N} \sum^N_{i=1} \big[ \log \sum_{a_j \in \mathcal{A}_i} \exp (U^i_{\theta_i} (s, a_{j})) - U^i_{\theta_i}(s, a_i) \big] \; ;
    \end{equation}
    where $a_{j}$ denotes the $j$th sub-action within the $i$th sub-action space, and $\alpha$ is a hyperparameter that controls the overall level of conservatism.  The full procedure can be found in Algorithm \ref{alg:DecQN+CQL}.  
    
    Note this particular implementation does not directly substitute the decomposition as per Equation \ref{Q-Decomp} into the original CQL loss.  Instead, we apply CQL directly to the utility values and then take the mean across sub-action dimensions.  This way we avoid having to estimate Q-values for all atomic actions (which a direct substitution would require) and we retain the equivalence to behavioural cloning at the sub-action level as
    \begin{equation*}
    \begin{split}
        \frac{\alpha}{|B|} \sum_{s, \mathbf{a} \sim \mathcal{B}} \frac{1}{N} \sum^N_{i=1} \log \sum_{a_j \in A_i} \exp (U^i_{\theta_i}(s,a_j)) - U^i_{\theta_i}(s,a_i) & =  - \frac{\alpha}{|B|} \sum_{s, \mathbf{a} \sim \mathcal{B}} \frac{1}{N} \sum^N_{i=1} \Big[ \log \frac{\exp (U^i_{\theta_i}(s,a_i))}{\sum_{a_j \in A_i} \exp (U^i_{\theta_i}(s,a_j))} \Big] \\
        & = - \frac{\alpha}{|B|} \sum_{s, \mathbf{a} \sim \mathcal{B}} \frac{1}{N} \sum^N_{i=1} \log \pi_{\theta_i}(a_i \mid s).
    \end{split}
    \end{equation*}    
    
\subsection{DecQN-IQL}
    Implicit Q-learning (IQL) \citep{kostrikov2021offlineImp} addresses the challenge of overestimation bias by learning a policy without having to query actions absent from data.  A state and state-action value function are trained on the data and then used to extract a policy via advantage-weighted-behavioural-cloning.

    The state value function $V_\psi(s)$ is trained via expectile regression, minimising the following loss:
    \[
    L(\psi) = \frac{1}{|B|} \sum_{s, a \sim \mathcal{B}} [ L^\tau_2 (Q_\theta(s, a) - V_\psi(s)) ] \; ;
    \]
    where if we denote $u=Q_\theta(s, a) - V_\psi(s)$ then $L^\tau_ 2(u)= | \tau - 1(u < 0) | u^2$ is the asymmetric least squares for the $\tau \in (0,1)$ expectile.  The state-action value function $Q_\theta(s, a)$ is trained using the same loss as Equation (\ref{BellmanOptimality}), with the target value now $y(r, s')=r + \gamma V_\psi(s')$.

    The policy follows that of discrete-action advantage-weighted-behavioural-cloning \citep{luo2023action}, which adapted to the factorised setting and integrated with DecQN leads to the following:
    \[
    \pi_i = \argmax_{a_i} \left[ \frac{1}{\lambda} A(s,a_i) + \log \pi^i_{\phi_i} (a_i \mid s) \right] \; .
    \]   
    where $A(s,a_i) = U(s,a_i) - V(s)$ is the (factorised) advantage function, $\pi^i_{\phi_i} (a_i \mid s)$ is the cloned behaviour policy for sub-action space $i$ trained via supervised learning and $\lambda$ is a hyperparameter controlling the balance between reinforcement learning and behavioural cloning.  The full procedure can be found in Algorithm \ref{alg:DecQN+IQL}.

\subsection{DecQN-OneStep}
    We can derive an alternative approach to IQL which removes the requirement for a separate state value function altogether.  Noting that $V(s) = \sum_a \pi(a \mid s) Q(s,a)$, we can instead use the cloned behaviour policy $\pi_\phi(a' \mid s')$ and state-action value function $Q_\theta(s',a')$ to calculate the state value function $V(s')$ instead.  This can be adapted and incorporated into DecQN by replacing $Q(s,a)$ with its decomposed form as per Equation (\ref{Q-Decomp}) and adjusting the policy to reflect a sub-action structure.  We denote this approach DecQN-OneStep as it mirrors one-step RL approaches that train state value functions using fitted Q-evaluation \citep{brandfonbrener2021offline}.  The full procedure can be found in Algorithm \ref{alg:DecQN+OneStep}.

\section{Environments and datasets}\label{EnvDatasets}
    At present, there are relatively few established environments/tasks specifically designed for factorised action spaces.  As such, there is an absence of benchmark datasets akin to those available for continuous or small-scale discrete action spaces such as D4RL \citep{fu2020d4rl} and RL Unplugged \citep{gulcehre2020rl}.  In light of this, we introduce our own benchmarking suite constituting the following environments and tasks.

\begin{figure}[ht]
            \centering
            \includegraphics[width=\linewidth]{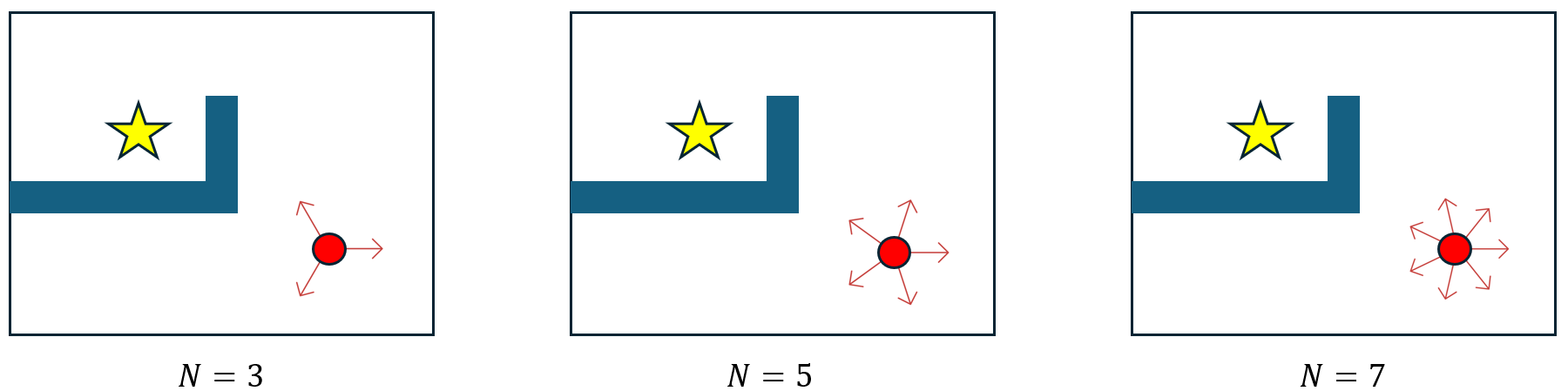}
            \caption{Examples of maze environment with different numbers of actuators.  The star represents the target goal location, the red dot the agent and the arrows the actuators.  Adapted from original Figure in \cite{chandak2019learning}.}
            \label{fig: MazeExamples}
        \end{figure}

    \textbf{Maze:} A simple maze-based environment first introduced by \cite{chandak2019learning} in which an agent is tasked with reaching a target goal location.  The state space is continuous and comprises the agent's current location in the maze.  The agent's movement is controlled by a series of $N$ actuators which can be turned on or off, with each actuator corresponding to a unit of movement in a single direction, with actuator $i$ applying force at an angle of $2\pi i /N$ when activated, as illustrated in Figure \ref{fig: MazeExamples}.  The net outcome of actuator selection is the vectoral summation of the movements associated with each actuator.  The agent's action space comprises the unique combination of the set of binary decision with respect to each actuator, leading to to an action set that is exponential in the number of actuators such that $|A| = 2^N$.  

    \textbf{DeepMind control suite:}  A discretised variant of the DeepMind control suite \citep{tunyasuvunakool2020dm_control}, as previously adopted by \cite{seyde2022solving, ireland2023revalued}.  This suite contains a variety of environments and tasks, which although originally designed for continuous control can easily be repurposed for a discrete factorisable setting by discretising individual sub-action spaces, thus creating scenarios that can vary in size and complexity.  This discretisation process involves selecting a subset of actions from the original continuous space which then become the discrete actions available to the agent.  For example, a continuous action in the range $[-1, 1]$ can be discretised into three discrete actions corresponding to the subset $\{-1, 0, 1\}$.  We emphasise this discretisation procedure happens prior to data collection, i.e. we are not discretising an existing continuous action dataset.

    The choice of environments and tasks reflects our desire to provide a benchmark that is amenable to factorisation and decomposition to varying degrees, allowing us to investigate the limits of the assumptions and conditions outlined in Section \ref{Case}.  For the maze task, the action set is inherently factorisable for each actuator, yet reaching the target goal requires at least some coordination across actuators .  For the DeepMind control suite, each sub-action corresponds to the adjustment of a specific robotic component (e.g. a joint), but to successfully complete the task may require alignment across some or all components.  We emphasise that our benchmark is not intended to cover large scale discrete action settings for which factorisation/decomposition is not feasible.

    For the datasets themselves, we follow a similar procedure to D4RL.  Using DecQN/REValueD, we train agents to ``expert'' and ``medium'' levels of performance and then collect transitions from the resulting policies.  Here, we define ``expert'' to be the peak performance achieved by DecQN/REValueD and ``medium'' to be approximately 1/3rd the performance of the ``expert''.  We create a third dataset ``medium-expert'' by combining transitions from these two sources and a fourth ``random-medium-expert'' containing transitions constituting 45\% random and medium transitions and 10\% expert.  Each of these datasets presents a specific challenge to agents, namely the ability to learn from optimal or sub-optimal data (``expert'' and ``medium'', respectively) as well as data that contains a mixture (``medium-expert'' and ``random-medium-expert'').  More details on this training and data collection procedure are provided in Appendix \ref{sec: data collection procedure}.

\section{Experimental evaluation}\label{experiments}
        We train agents using DecQN, DecQN-BCQ, DecQN-CQL, DecQN-IQL and DecQN-OneStep on our benchmark datasets and evaluate their performance in the simulated environment.  We also train and evaluate agents using a factorised equivalent of behavioural cloning to provide a supervised learning baseline.  Performance is measured in terms of normalised score, where $score_{norm} = 100 \times \frac{score - score_{random}}{score_{expert} - score_{random}}$  with 0 representing a random policy and 100 the ``expert'' policy from the fully trained agent.  We repeat experiments across five random seeds, reporting results as mean normalised scores $\pm$ one standard error across seeds.  For each set of experiments we provide visual summaries with tabulated results available in Appendix \ref{App_Tab_Results} for completeness.  Full implementation details, including network architectures, hyperparameters and training procedures are provided in Appendix \ref{appendix:full_imp}  

    \subsection{Case study: DQN-CQL vs DecQN-CQL}\label{AtomicVsDecQN}    
        Before considering the full benchmark, we conduct a case study directly comparing  DQN-CQL (i.e. CQL under atomic action representation) and DecQN-CQL.  Following the procedure outlined in Section \ref{EnvDatasets} we construct a ``medium-expert'' dataset for the ``cheetah-run'' task for number of bins $n_i \in \{3, 4, 5, 6\}$ and compare the performance of resulting policies and overall computation time.  In addition we construct a ``random-medium-expert'' dataset for the Maze task with $N=15$ actuators for varying dataset sizes and compare performance and computational resources.
        
\begin{figure}[ht]
                \centering
                \includegraphics[width=\linewidth]{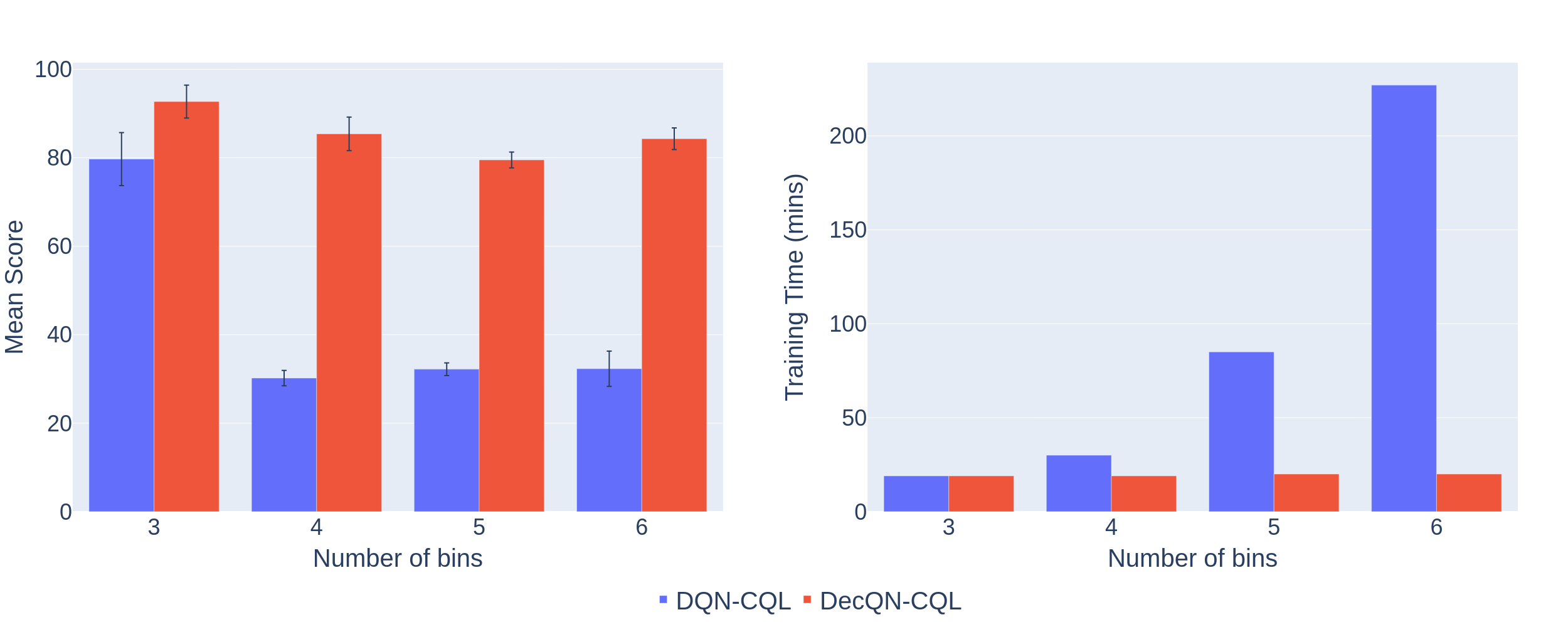}
                \caption{Comparisons of performance (left) and computation time (right) for DQN-CQL and DecQN-CQL on the ``cheetah-run-medium-expert'' dataset for varying numbers of bins.  As the number of bins increases, DQN-CQL suffers notable drops in performance and increases in computation time, whereas DecQN-CQL is relatively resilient in both areas.}
                \label{fig: FactVsAtomic}
            \end{figure}

        In Figure  \ref{fig: FactVsAtomic} we see that as the number of sub-actions $n_i$ increases for the ``cheetah-run'' task, DQN-CQL exhibits a notable decline in performance, whereas DecQN-CQL performance declines only marginally.  We also see a dramatic increase in computation time for DQN-CQL whereas DecQN-CQL remains roughly constant.  For DQN-CQL, these outcomes are symptomatic of the combinatorial explosion in the number of actions requiring value-estimation and the associated number of out-of-distribution global actions.  These issues are less prevalent in DecQN-CQL due to its factorised and decomposed formulation.  To provide further insights, in Appendix \ref{CaseStudyErrors} we also examine the evolution of Q-values errors during training, finding these errors are consistently lower for DecQN-CQL than DQN-CQL, aligning with each algorithm's respective performance.

\begin{figure}[ht]
                \centering
                \includegraphics[width=\linewidth]{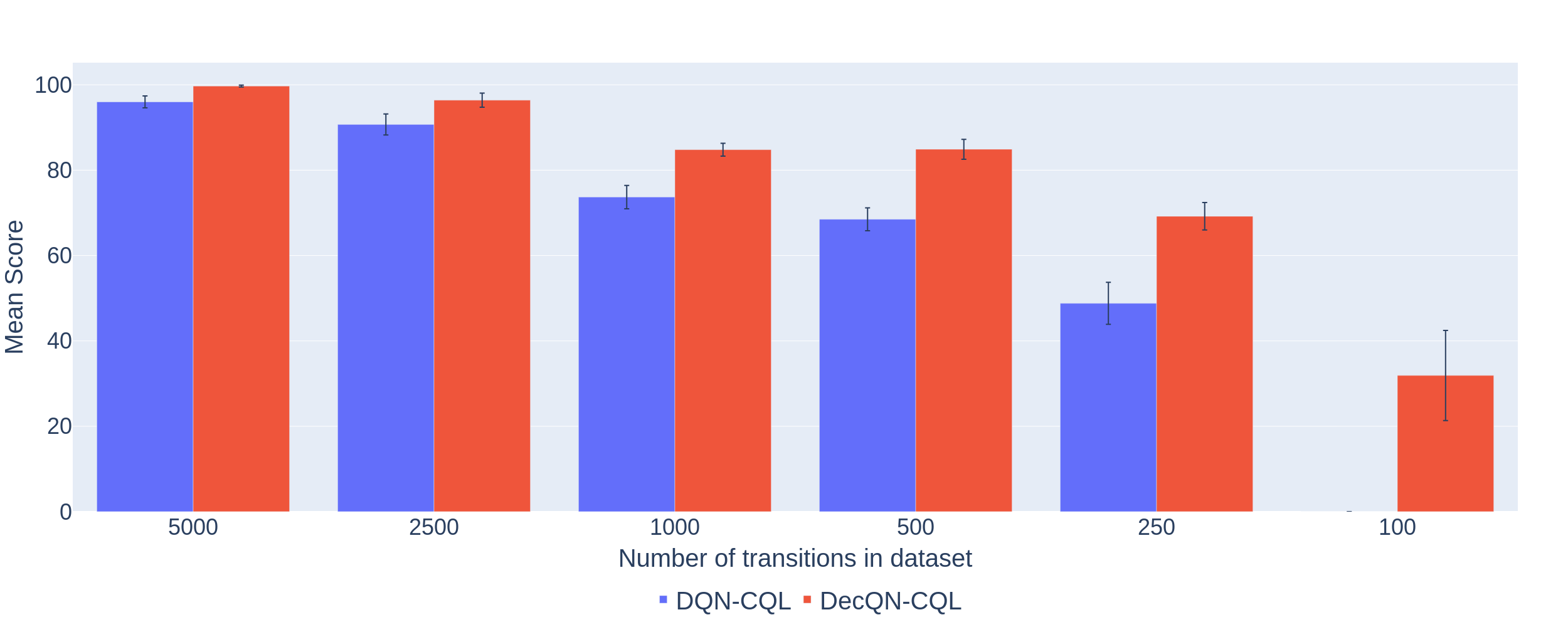}
                \caption{Comparison of performance for DQN-CQL and DecQN-CQL on the Maze task with $N=15$ actuators with ``random-medium-expert'' datasets of varying size.  As the number of transitions in the dataset decreases, DQN-CQL suffers more notable drops in performance in comparison to DecQN-CQL.}
                \label{fig: FactVsAtomicMaze}
            \end{figure}

        In Figure \ref{fig: FactVsAtomicMaze} we see that as the size of the dataset for the Maze task decreases, DQN-CQL exhibits a more notable decline in performance than DecQN-CQL.  This is particularly the case when the size of the dataset is very small, i.e. $\le$ 250 transitions.  These outcomes support our notion of better action coverage under a factorised representation than atomic, where value estimates for observed sub-actions can be combined to provide better estimates of global actions that haven't been observed.  Furthermore, this is achieved through much more computationally efficient means, with training time for DecQN-CQL being over 8 times faster than DQN-CQL (4mins vs 34mins, respectively) and GPU usage 7 times less (246MBs vs 1728MBs, respectively).

    \subsection{Maze}
\begin{figure}[ht]
            \centering
            \includegraphics[width=\linewidth]{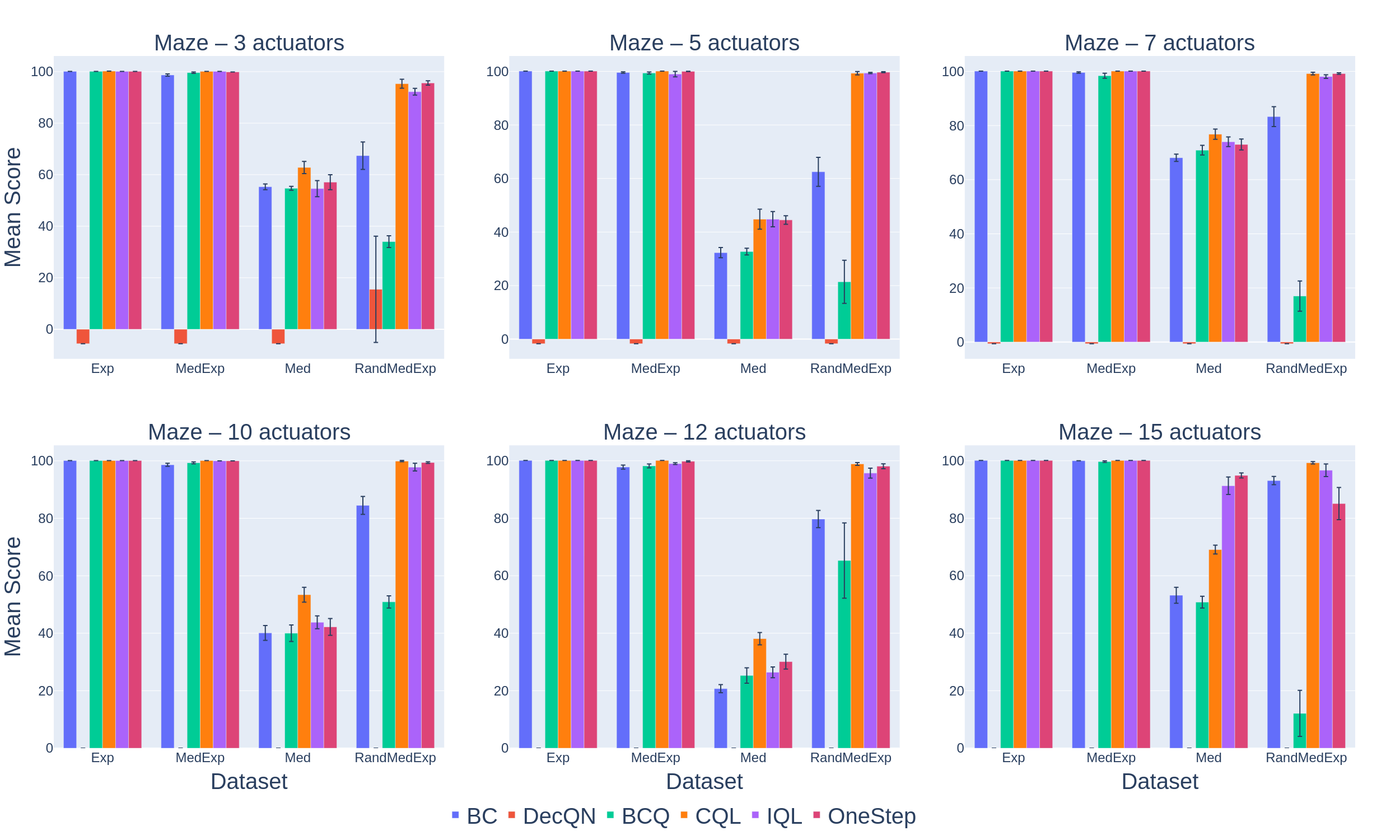}
            \caption{Performance comparison on maze task for varying numbers of actuators.  For presentation purposes the prefix ``DecQN-'' has been omitted for each offline method.  In general, all offline RL methods improve over behavioural cloning, with the exception of DecQN-BCQ for ``random-medium-expert'' datasets.  DecQN without any offline modification performs poorly across the board. \label{Fig: MazeResults}}
        \end{figure}

    In Figure \ref{Fig: MazeResults} we summarise each algorithm's performance for varying numbers of actuators and dataset composition.  Across the board, we see that DecQN sans offline modification results in policies that perform no better than random, a direct consequence of aforementioned issues relating to overestimation bias.  With the exception of DecQN-BCQ for ``random-medium-expert'' datasets, all offline methods match or outperform behavioural cloning, particularly for datasets constituting high levels of sub-optimal trajectories (``medium'' and ``random-medium-expert'').  For datasets containing a proportion of expert trajectories (``medium-expert'' and ``random-medium-expert'') all offline methods (again with the exception of DecQN-BCQ) are able extract expert or near-expert level policies.  \footnote{For `medium-expert'' datasets we note that BC is able to achieve expert-level performance on par with our offline methods.  We attribute this to the relative simplicity of the environment coupled with a reasonable level of expert-level trajectories within the datasets.  Such outcomes are not replicated in DMC tasks which are more complex and have much longer trajectories}  Comparing just the offline RL methods, we see there is little to separate them for ``expert'' and ``medium-expert'' regardless of the number of actuators.  For ``medium'' and ``random-medium-expert'', DecQN-CQL exhibits an edge when the number of actuators is between $3$ and $12$, but for ``medium''  this is lost to DecQN-IQL/OneStep as the number of actuators increases to $15$.

    \subsection{DeepMind Control Suite}\label{main_results}
\begin{figure}[ht]
            \centering
            \includegraphics[width=\linewidth]{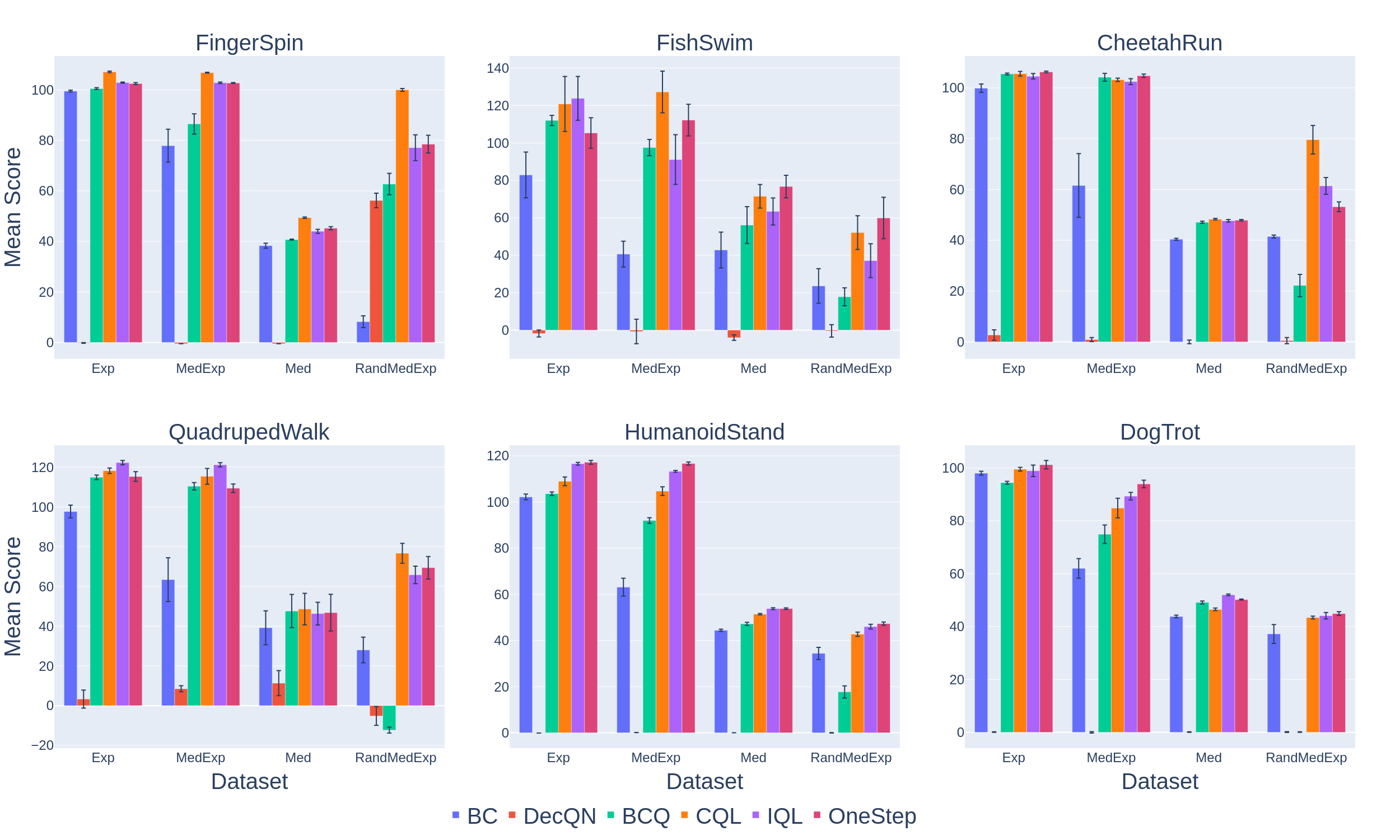}
            \caption{Performance comparison across benchmark for $n_i=3$.  In general, all offline RL methods improve over behavioural cloning, with the exception of DecQN-BCQ for ``random-medium-expert'' datasets.  DecQN without any offline modification performs poorly across all environments/tasks. \label{Fig: MainResults}}
        \end{figure}

    In Figure \ref{Fig: MainResults} we summarise each algorithm's performance across the full range of tasks, setting the number of sub-actions $n_i = 3$.  This necessitates the use of value-decomposition for all but the most simple tasks, as highlighted in Table \ref{tab:dm_control_suite}.  In general we see that all offline RL methods outperform behavioural cloning across all environments/tasks and datasets, with the exception of DecQN-BCQ for ``random-medium-expert'' datasets which performs quite poorly.  In terms of offline methods specifically, in general DecQN-CQL has a slight edge over others for lower dimensional tasks such as ``finger-spin'', ``fish-swim'' and ``cheetah-run'', whilst DecQN-IQL/OneStep have the edge for higher-dimensional tasks such as ``humanoid-stand'' and ``dog-trot''.  For ``medium-expert'' datasets we see in most cases all offline methods are able to learn expert or near-expert level policies.  Extracting optimal behaviour from ``random-medium-expert'' datasets proves significantly more challenging than in the Maze environment, likely a result of these environments/tasks being much more complex coupled with datasets being both highly variable and constituting relatively few expert trajectories in relation to trajectory length.

    \subsubsection{Increasing the number of sub-actions}
        To help provide insights into the ability of our chosen offline methods to scale to larger and larger action spaces, we increase the number of sub-actions within each sub-action space and repeat our experiments.  We focus in particularly on the dog-trot environment since this is by far the largest in terms of actions.  We collect datasets following the same procedure outlined in Section \ref{EnvDatasets} for number of bins $n_i\in\{10, 30, 50, 75, 100\}$.  
        
\begin{figure}[ht]
                \centering
                \includegraphics[width=\linewidth]{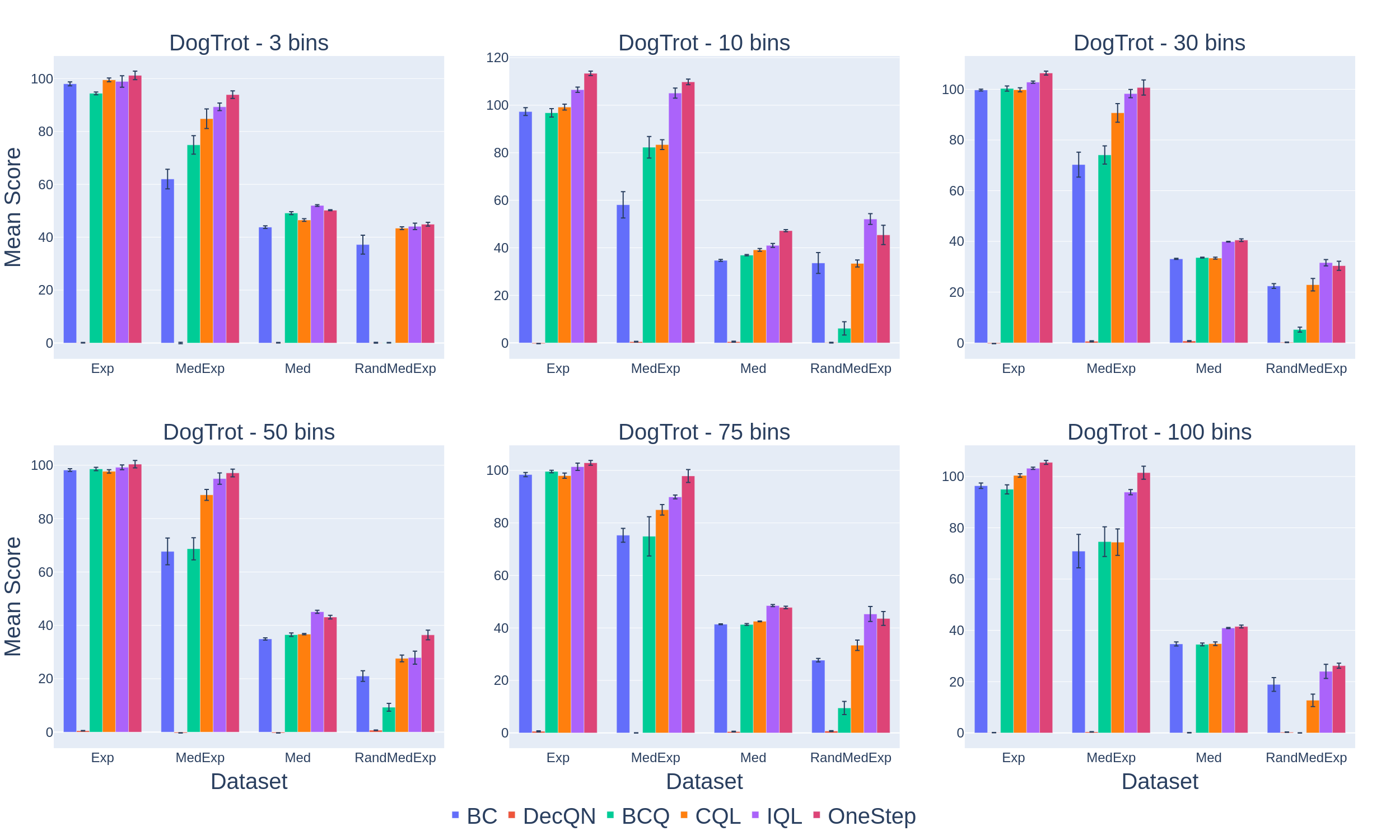}
                \caption{Performance comparison for dog-trot for $n_i\in\{3, 10, 30, 50, 75, 100\}$.  Each approach is reasonably resilient to increases in the number of bins, although for ``random-medium-expert'' datasets extracting a good policy appears to become more challenging as $n$ gets very large. \label{Fig: LargerBinResults}}
            \end{figure}

        We summarise results in Figure \ref{Fig: LargerBinResults}.  In general, we see that our chosen offline methods are robust to increases in the number of bins, continuing to outperform behavioural cloning (with the same exception for DecQN-BCQ on ``random-medium-expert'') and extract near-expert policies from ``medium-expert'' datasets, with DecQN-IQL/-OneStep maintaining their edge over DecQN-CQL.  For ``random-medium-expert'' datasets we start to notice a decline in performance as we approach the upper end of our number of bins, most noticeably when $n=100$.  This is likely a consequence of more and more actions exacerbating the difficulties in obtaining good policies from highly variable and largely sub-optimal data.
        
\section{Discussion and conclusion}\label{DiscConc}
    In this work, we have conducted a formative investigation of offline reinforcement in factorisable action spaces.  Through empirical evaluation, we have shown how a factorised and decomposed approach offers numerous benefits over standard/atomic approaches.  Using a bespoke benchmark we have undertaken an extensive empirical evaluation of several offline RL approaches adapted to this factorised and decomposed framework, providing insights into each approach's ability to learn tasks of varying complexity from datasets of differing size and quality.  
    
    In general, our empirical evaluation demonstrates our chosen offline methods adapt well to the factorised setting when combined with value-decomposition in the form of DecQN.  With one exception, all approaches are consistently able to outperform behavioural cloning regardless of data quality, and where datasets contain sufficient levels of high-quality trajectories (i.e. ``expert'' and ``medium-expert''), obtain expert/near-expert policies.  There is however notable room for improvement for datasets with a scarcity of high-quality trajectories for more complex tasks (i.e. ``medium'' and ``random-medium-expert'' for DMC environments/tasks).

    Our initial investigation opens up numerous other possibilities for future research.  One of these is the development of techniques for automatically tuning hyperparameters during training, which at present are not environment/task agnostic (only data quality).  In addition, as with continuous counterparts, performance can be enhanced by allowing hyperparameters to vary for each dataset (see Appendix \ref{sec: allowing variation within task} for examples).  Off-policy evaluation could also prove beneficial here \citep{rebello2023leveraging}, providing assurances on the quality of a policy prior to deployment.  
    
    For DecQN-BCQ/-IQL/-OneStep, alternative approaches to modelling the behaviour policy $\pi_\phi$ may help improve performance for more challenging datasets. Incorporating other methods outlined in Section \ref{RelatedWork} may also prove beneficial.  For example, the use of ensembles for capturing uncertainty in value estimates has been shown to perform well in combination with behavioural cloning in continuous action spaces \citep{beeson2024balancing}, and is a relatively straightforward extension to the approaches we consider here.
    
    Whilst DecQN offers a simple, effective and computationally efficient foundation for offline RL in factorisable action spaces, we note there are some inherent assumptions and limitations to the value decomposition approach that warrant further investigation, as noted in Sections \ref{Case: Approp} - \ref{Case: Limit}.  In particular, the efficacy of DecQN relies on individual sub-action optimisation leading to globally optimal joint policies. However, for tasks with sparse rewards or complex sub-action dependencies, individually learned sub-policies may fail to properly compose into a coherent overall policy. For example, in assembly tasks, separately learned pick, place, and connect skills could lead to conflicting behaviors when combined. Additional research into modeling sub-action interactions during decomposition could help overcome this limitation.

    Our technical analysis can also potentially be expanded upon by considering the degree to which a global action is in-distribution/out-of-distribution.  Implicitly we have assumed errors from actions/sub-actions are the same regardless of how many times a global action appears in the dataset.  Relaxing this assumption could help refine the analysis, particularly for stochastic environments where multiple interactions can provide additional information.  Further refinement may also be possible by addressing assumptions on Q-function decomposition (as highlighted in Sections \ref{Case: Approp} - \ref{Case: Limit}), particularly the true Q-function which may not necessarily decompose in the same manner as DecQN. 

    Beyond algorithmic considerations, there are also several avenues for exploration in relation to the environments and tasks themselves.  While our benchmark provides a broad range of diverse tasks, any particular one is deficient in either complexity (i.e. maze) or being naturally factorisable (i.e. DMC).  Thus, the creation of bespoke environments that help bridge this gap would be a valuable contribution, providing more realistic scenarios to evaluate against.  In addition, future work could investigate more nuanced aspects relating to action space discretisation, such as variable numbers of bins, non-even spacing between actions, clustering of actions and masked actions.

    We hope our work underscores the unique setting and challenges of conducting offline RL in factorisable action spaces and paves the way for future research by providing an accessible and solid foundation from which to build upon.

\textbf{Acknowledgments}
\noindent
AB acknowledges support from University of Warwick and University of Birmingham NHS Foundation Trust. 
\noindent
GM acknowledges support from a UKRI AI Turing Acceleration Fellowship (EPSRC EP/V024868/1). 

\newpage
\bibliography{main}

\begin{thebibliography}{71}
\providecommand{\natexlab}[1]{#1}
\providecommand{\url}[1]{\texttt{#1}}
\expandafter\ifx\csname urlstyle\endcsname\relax
  \providecommand{\doi}[1]{doi: #1}\else
  \providecommand{\doi}{doi: \begingroup \urlstyle{rm}\Url}\fi

\bibitem[An et~al.(2021)An, Moon, Kim, and Song]{an2021uncertainty}
Gaon An, Seungyong Moon, Jang-Hyun Kim, and Hyun~Oh Song.
\newblock Uncertainty-based offline reinforcement learning with diversified {Q-ensemble}.
\newblock \emph{Advances in neural information processing systems}, 34:\penalty0 7436--7447, 2021.

\bibitem[Argenson and Dulac-Arnold(2020)]{argenson2020model}
Arthur Argenson and Gabriel Dulac-Arnold.
\newblock Model-based offline planning.
\newblock In \emph{International Conference on Learning Representations}, 2020.

\bibitem[Bai et~al.(2022)Bai, Wang, Yang, Deng, Garg, Liu, and Wang]{baipessimistic}
Chenjia Bai, Lingxiao Wang, Zhuoran Yang, Zhi-Hong Deng, Animesh Garg, Peng Liu, and Zhaoran Wang.
\newblock Pessimistic bootstrapping for uncertainty-driven offline reinforcement learning.
\newblock In \emph{International Conference on Learning Representations}, 2022.

\bibitem[Beeson and Montana(2024)]{beeson2024balancing}
Alex Beeson and Giovanni Montana.
\newblock Balancing policy constraint and ensemble size in uncertainty-based offline reinforcement learning.
\newblock \emph{Machine Learning}, 113\penalty0 (1):\penalty0 443--488, 2024.

\bibitem[Brandfonbrener et~al.(2021)Brandfonbrener, Whitney, Ranganath, and Bruna]{brandfonbrener2021offline}
David Brandfonbrener, Will Whitney, Rajesh Ranganath, and Joan Bruna.
\newblock Offline {RL} without off-policy evaluation.
\newblock \emph{Advances in neural information processing systems}, 34:\penalty0 4933--4946, 2021.

\bibitem[Chandak et~al.(2019)Chandak, Theocharous, Kostas, Jordan, and Thomas]{chandak2019learning}
Yash Chandak, Georgios Theocharous, James Kostas, Scott Jordan, and Philip Thomas.
\newblock Learning action representations for reinforcement learning.
\newblock In \emph{International conference on machine learning}, pages 941--950. PMLR, 2019.

\bibitem[Chebotar et~al.(2023)Chebotar, Vuong, Hausman, Xia, Lu, Irpan, Kumar, Yu, Herzog, Pertsch, et~al.]{chebotar2023q}
Yevgen Chebotar, Quan Vuong, Karol Hausman, Fei Xia, Yao Lu, Alex Irpan, Aviral Kumar, Tianhe Yu, Alexander Herzog, Karl Pertsch, et~al.
\newblock Q-transformer: Scalable offline reinforcement learning via autoregressive {Q-functions}.
\newblock In \emph{Conference on Robot Learning}, pages 3909--3928. PMLR, 2023.

\bibitem[Claus and Boutilier(1998)]{claus1998dynamics}
Caroline Claus and Craig Boutilier.
\newblock The dynamics of reinforcement learning in cooperative multiagent systems.
\newblock \emph{AAAI/IAAI}, 1998\penalty0 (746-752):\penalty0 2, 1998.

\bibitem[Driess et~al.(2020)Driess, Ha, and Toussaint]{driess2020deep}
Driess, Ha, and Toussaint.
\newblock Deep visual reasoning - learning to predict action sequences for assembly tasks.
\newblock In \emph{2020 IEEE/RSJ International Conference on Intelligent Robots and Systems (IROS)}, pages 4645--4651. IEEE, 2020.

\bibitem[Du et~al.(2022)Du, Ding, Guo, Zhang, Zhang, and Ding]{du2022value}
Wei Du, Shifei Ding, Lili Guo, Jian Zhang, Chenglong Zhang, and Ling Ding.
\newblock Value function factorization with dynamic weighting for deep multi-agent reinforcement learning.
\newblock \emph{Information Sciences}, 615:\penalty0 191--208, 2022.

\bibitem[Dulac-Arnold et~al.(2015)Dulac-Arnold, Evans, van Hasselt, Sunehag, Lillicrap, Hunt, Mann, Weber, Degris, and Coppin]{dulac2015deep}
Gabriel Dulac-Arnold, Richard Evans, Hado van Hasselt, Peter Sunehag, Timothy Lillicrap, Jonathan Hunt, Timothy Mann, Theophane Weber, Thomas Degris, and Ben Coppin.
\newblock Deep reinforcement learning in large discrete action spaces.
\newblock \emph{arXiv preprint arXiv:1512.07679}, 2015.

\bibitem[Farquhar et~al.(2020)Farquhar, Gustafson, Lin, Whiteson, Usunier, and Synnaeve]{farquhar2020growing}
Gregory Farquhar, Laura Gustafson, Zeming Lin, Shimon Whiteson, Nicolas Usunier, and Gabriel Synnaeve.
\newblock Growing action spaces.
\newblock In \emph{International Conference on Machine Learning}, pages 3040--3051. PMLR, 2020.

\bibitem[Fu et~al.(2020)Fu, Kumar, Nachum, Tucker, and Levine]{fu2020d4rl}
Justin Fu, Aviral Kumar, Ofir Nachum, George Tucker, and Sergey Levine.
\newblock {D4RL}: Datasets for deep data-driven reinforcement learning.
\newblock \emph{arXiv preprint arXiv:2004.07219}, 2020.

\bibitem[Fujimoto and Gu(2021)]{fujimoto2021minimalist}
Scott Fujimoto and Shixiang~Shane Gu.
\newblock A minimalist approach to offline reinforcement learning.
\newblock \emph{Advances in Neural Information Processing Systems}, 34:\penalty0 20132--20145, 2021.

\bibitem[Fujimoto et~al.(2018)Fujimoto, Hoof, and Meger]{fujimoto2018addressing}
Scott Fujimoto, Herke Hoof, and David Meger.
\newblock Addressing function approximation error in actor-critic methods.
\newblock In \emph{International conference on machine learning}, pages 1587--1596. PMLR, 2018.

\bibitem[Fujimoto et~al.(2019{\natexlab{a}})Fujimoto, Conti, Ghavamzadeh, and Pineau]{fujimoto2019benchmarking}
Scott Fujimoto, Edoardo Conti, Mohammad Ghavamzadeh, and Joelle Pineau.
\newblock Benchmarking batch deep reinforcement learning algorithms.
\newblock \emph{arXiv preprint arXiv:1910.01708}, 2019{\natexlab{a}}.

\bibitem[Fujimoto et~al.(2019{\natexlab{b}})Fujimoto, Meger, and Precup]{fujimoto2019off}
Scott Fujimoto, David Meger, and Doina Precup.
\newblock Off-policy deep reinforcement learning without exploration.
\newblock In \emph{International Conference on Machine Learning}, pages 2052--2062. PMLR, 2019{\natexlab{b}}.

\bibitem[Gaskett(2002)]{gaskett2002q}
Chris Gaskett.
\newblock \emph{Q-learning for robot control}.
\newblock PhD thesis, Australian National University, 2002.

\bibitem[Ghasemipour et~al.(2022)Ghasemipour, Gu, and Nachum]{ghasemipour2022so}
Kamyar Ghasemipour, Shixiang~Shane Gu, and Ofir Nachum.
\newblock Why so pessimistic? estimating uncertainties for offline {RL} through ensembles, and why their independence matters.
\newblock \emph{Advances in Neural Information Processing Systems}, 35:\penalty0 18267--18281, 2022.

\bibitem[Gottesman et~al.(2018)Gottesman, Johansson, Meier, Dent, Lee, Srinivasan, Zhang, Ding, Wihl, Peng, et~al.]{gottesman2018evaluating}
Omer Gottesman, Fredrik Johansson, Joshua Meier, Jack Dent, Donghun Lee, Srivatsan Srinivasan, Linying Zhang, Yi~Ding, David Wihl, Xuefeng Peng, et~al.
\newblock Evaluating reinforcement learning algorithms in observational health settings.
\newblock \emph{arXiv preprint arXiv:1805.12298}, 2018.

\bibitem[Gu et~al.(2022)Gu, Zhao, Chen, Li, Hao, and An]{gu2022learning}
Pengjie Gu, Mengchen Zhao, Chen Chen, Dong Li, Jianye Hao, and Bo~An.
\newblock Learning pseudometric-based action representations for offline reinforcement learning.
\newblock In \emph{International Conference on Machine Learning}, pages 7902--7918. PMLR, 2022.

\bibitem[Guestrin et~al.(2003)Guestrin, Koller, Parr, and Venkataraman]{guestrin2003efficient}
Carlos Guestrin, Daphne Koller, Ronald Parr, and Shobha Venkataraman.
\newblock Efficient solution algorithms for factored {MDPs}.
\newblock \emph{Journal of Artificial Intelligence Research}, 19:\penalty0 399--468, 2003.

\bibitem[Gulcehre et~al.(2020{\natexlab{a}})Gulcehre, Colmenarejo, Sygnowski, Paine, Zolna, Chen, Hoffman, Pascanu, de~Freitas, et~al.]{gulcehre2020addressing}
Caglar Gulcehre, Sergio~G{\'o}mez Colmenarejo, Jakub Sygnowski, Thomas Paine, Konrad Zolna, Yutian Chen, Matthew Hoffman, Razvan Pascanu, Nando de~Freitas, et~al.
\newblock Addressing extrapolation error in deep offline reinforcement learning.
\newblock \emph{Offline Reinforcement Learning Workshop at Neural Information Processing Systems, 2020}, 2020{\natexlab{a}}.

\bibitem[Gulcehre et~al.(2020{\natexlab{b}})Gulcehre, Wang, Novikov, Paine, G{\'o}mez, Zolna, Agarwal, Merel, Mankowitz, Paduraru, et~al.]{gulcehre2020rl}
Caglar Gulcehre, Ziyu Wang, Alexander Novikov, Thomas Paine, Sergio G{\'o}mez, Konrad Zolna, Rishabh Agarwal, Josh~S Merel, Daniel~J Mankowitz, Cosmin Paduraru, et~al.
\newblock Rl unplugged: A suite of benchmarks for offline reinforcement learning.
\newblock \emph{Advances in Neural Information Processing Systems}, 33:\penalty0 7248--7259, 2020{\natexlab{b}}.

\bibitem[Hessel et~al.(2018)Hessel, Modayil, Van~Hasselt, Schaul, Ostrovski, Dabney, Horgan, Piot, Azar, and Silver]{hessel2018rainbow}
Matteo Hessel, Joseph Modayil, Hado Van~Hasselt, Tom Schaul, Georg Ostrovski, Will Dabney, Dan Horgan, Bilal Piot, Mohammad Azar, and David Silver.
\newblock Rainbow: Combining improvements in deep reinforcement learning.
\newblock In \emph{Thirty-second AAAI conference on artificial intelligence}, 2018.

\bibitem[Horgan et~al.(2018)Horgan, Quan, Budden, Barth-Maron, Hessel, van Hasselt, and Silver]{horgan2018distributed}
Dan Horgan, John Quan, David Budden, Gabriel Barth-Maron, Matteo Hessel, Hado van Hasselt, and David Silver.
\newblock Distributed prioritized experience replay.
\newblock In \emph{International Conference on Learning Representations}, 2018.

\bibitem[Hubert et~al.(2021)Hubert, Schrittwieser, Antonoglou, Barekatain, Schmitt, and Silver]{hubert2021learning}
Thomas Hubert, Julian Schrittwieser, Ioannis Antonoglou, Mohammadamin Barekatain, Simon Schmitt, and David Silver.
\newblock Learning and planning in complex action spaces.
\newblock In \emph{International Conference on Machine Learning}, pages 4476--4486. PMLR, 2021.

\bibitem[Ireland and Montana(2023)]{ireland2023revalued}
David Ireland and Giovanni Montana.
\newblock Revalued: Regularised ensemble value-decomposition for factorisable {Markov} decision processes.
\newblock In \emph{The Twelfth International Conference on Learning Representations}, 2023.

\bibitem[Janner et~al.(2022)Janner, Du, Tenenbaum, and Levine]{janner2022planning}
Michael Janner, Yilun Du, Joshua Tenenbaum, and Sergey Levine.
\newblock Planning with diffusion for flexible behavior synthesis.
\newblock In \emph{International Conference on Machine Learning}, pages 9902--9915. PMLR, 2022.

\bibitem[Kalashnikov et~al.(2018)Kalashnikov, Irpan, Pastor, Ibarz, Herzog, Jang, Quillen, Holly, Kalakrishnan, Vanhoucke, et~al.]{kalashnikov2018scalable}
Dmitry Kalashnikov, Alex Irpan, Peter Pastor, Julian Ibarz, Alexander Herzog, Eric Jang, Deirdre Quillen, Ethan Holly, Mrinal Kalakrishnan, Vincent Vanhoucke, et~al.
\newblock Scalable deep reinforcement learning for vision-based robotic manipulation.
\newblock In \emph{Conference on robot learning}, pages 651--673. PMLR, 2018.

\bibitem[Kearns and Koller(1999)]{kearns1999efficient}
Michael Kearns and Daphne Koller.
\newblock Efficient reinforcement learning in factored {MDPs}.
\newblock In \emph{IJCAI}, volume~16, pages 740--747, 1999.

\bibitem[Kidambi et~al.(2020)Kidambi, Rajeswaran, Netrapalli, and Joachims]{kidambi2020morel}
Rahul Kidambi, Aravind Rajeswaran, Praneeth Netrapalli, and Thorsten Joachims.
\newblock {MOReL}: Model-based offline reinforcement learning.
\newblock \emph{Advances in neural information processing systems}, 33:\penalty0 21810--21823, 2020.

\bibitem[Kingma and Ba(2014)]{kingma2014adam}
Diederik~P Kingma and Jimmy Ba.
\newblock Adam: A method for stochastic optimization.
\newblock \emph{arXiv preprint arXiv:1412.6980}, 2014.

\bibitem[Kiran et~al.(2022)Kiran, Sobh, Talpaert, Mannion, Sallab, Yogamani, and Pérez]{RLAutoDriving}
B~Ravi Kiran, Ibrahim Sobh, Victor Talpaert, Patrick Mannion, Ahmad A.~Al Sallab, Senthil Yogamani, and Patrick Pérez.
\newblock Deep reinforcement learning for autonomous driving: A survey.
\newblock \emph{IEEE Transactions on Intelligent Transportation Systems}, 23\penalty0 (6):\penalty0 4909--4926, 2022.

\bibitem[Kostrikov et~al.(2021{\natexlab{a}})Kostrikov, Fergus, Tompson, and Nachum]{kostrikov2021offline}
Ilya Kostrikov, Rob Fergus, Jonathan Tompson, and Ofir Nachum.
\newblock Offline reinforcement learning with {Fisher} divergence critic regularization.
\newblock In \emph{International Conference on Machine Learning}, pages 5774--5783. PMLR, 2021{\natexlab{a}}.

\bibitem[Kostrikov et~al.(2021{\natexlab{b}})Kostrikov, Nair, and Levine]{kostrikov2021offlineImp}
Ilya Kostrikov, Ashvin Nair, and Sergey Levine.
\newblock Offline reinforcement learning with implicit {Q-Learning}.
\newblock In \emph{International Conference on Learning Representations}, 2021{\natexlab{b}}.

\bibitem[Kraemer and Banerjee(2016)]{kraemer2016multi}
Landon Kraemer and Bikramjit Banerjee.
\newblock Multi-agent reinforcement learning as a rehearsal for decentralized planning.
\newblock \emph{Neurocomputing}, 190:\penalty0 82--94, 2016.

\bibitem[Kumar et~al.(2019)Kumar, Fu, Soh, Tucker, and Levine]{kumar2019stabilizing}
Aviral Kumar, Justin Fu, Matthew Soh, George Tucker, and Sergey Levine.
\newblock Stabilizing off-policy {Q-learning} via bootstrapping error reduction.
\newblock \emph{Advances in neural information processing systems}, 32, 2019.

\bibitem[Kumar et~al.(2020)Kumar, Zhou, Tucker, and Levine]{kumar2020conservative}
Aviral Kumar, Aurick Zhou, George Tucker, and Sergey Levine.
\newblock Conservative {Q-learning} for offline reinforcement learning.
\newblock \emph{Advances in Neural Information Processing Systems}, 33:\penalty0 1179--1191, 2020.

\bibitem[Lange et~al.(2012)Lange, Gabel, and Riedmiller]{lange2012batch}
Sascha Lange, Thomas Gabel, and Martin Riedmiller.
\newblock Batch reinforcement learning.
\newblock \emph{Springer}, pages 45--73, 2012.

\bibitem[Levine et~al.(2020)Levine, Kumar, Tucker, and Fu]{levine2020offline}
Sergey Levine, Aviral Kumar, George Tucker, and Justin Fu.
\newblock Offline reinforcement learning: Tutorial, review, and perspectives on open problems.
\newblock \emph{arXiv preprint arXiv:2005.01643}, 2020.

\bibitem[Lin et~al.(2018)Lin, Zhao, Xu, and Zhou]{lin2018efficient}
Kaixiang Lin, Renyu Zhao, Zhe Xu, and Jiayu Zhou.
\newblock Efficient large-scale fleet management via multi-agent deep reinforcement learning.
\newblock In \emph{Proceedings of the 24th ACM SIGKDD international conference on knowledge discovery \& data mining}, pages 1774--1783, 2018.

\bibitem[Liu et~al.(2020)Liu, See, Ngiam, Celi, Sun, and Feng]{liu2020reinforcement}
Siqi Liu, Kay~Choong See, Kee~Yuan Ngiam, Leo~Anthony Celi, Xingzhi Sun, and Mengling Feng.
\newblock Reinforcement learning for clinical decision support in critical care: comprehensive review.
\newblock \emph{Journal of medical Internet research}, 22\penalty0 (7):\penalty0 e18477, 2020.

\bibitem[Luo et~al.(2023)Luo, Dong, Wu, Kumar, Geng, and Levine]{luo2023action}
Jianlan Luo, Perry Dong, Jeffrey Wu, Aviral Kumar, Xinyang Geng, and Sergey Levine.
\newblock Action-quantized offline reinforcement learning for robotic skill learning.
\newblock In \emph{7th Annual Conference on Robot Learning}, 2023.

\bibitem[Mahmood et~al.(2018)Mahmood, Korenkevych, Vasan, Ma, and Bergstra]{mahmood2018benchmarking}
A~Rupam Mahmood, Dmytro Korenkevych, Gautham Vasan, William Ma, and James Bergstra.
\newblock Benchmarking reinforcement learning algorithms on real-world robots.
\newblock In \emph{Conference on robot learning}, pages 561--591. PMLR, 2018.

\bibitem[Metz et~al.(2017)Metz, Ibarz, Jaitly, and Davidson]{metz2017discrete}
Luke Metz, Julian Ibarz, Navdeep Jaitly, and James Davidson.
\newblock Discrete sequential prediction of continuous actions for deep {RL}.
\newblock \emph{arXiv preprint arXiv:1705.05035}, 2017.

\bibitem[Mnih et~al.(2013)Mnih, Kavukcuoglu, Silver, Graves, Antonoglou, Wierstra, and Riedmiller]{mnih2013playing}
Volodymyr Mnih, Koray Kavukcuoglu, David Silver, Alex Graves, Ioannis Antonoglou, Daan Wierstra, and Martin Riedmiller.
\newblock Playing atari with deep reinforcement learning.
\newblock \emph{arXiv preprint arXiv:1312.5602}, 2013.

\bibitem[Nikulin et~al.(2023)Nikulin, Kurenkov, Tarasov, and Kolesnikov]{nikulin2023anti}
Alexander Nikulin, Vladislav Kurenkov, Denis Tarasov, and Sergey Kolesnikov.
\newblock Anti-exploration by random network distillation.
\newblock In \emph{International Conference on Machine Learning}, pages 26228--26244. PMLR, 2023.

\bibitem[Pierrot et~al.(2021)Pierrot, Mac{\'e}, Sevestre, Monier, Laterre, Perrin, Beguir, and Sigaud]{pierrotfactored}
Thomas Pierrot, Valentin Mac{\'e}, Jean-Baptiste Sevestre, Louis Monier, Alexandre Laterre, Nicolas Perrin, Karim Beguir, and Olivier Sigaud.
\newblock Factored action spaces in deep reinforcement learning.
\newblock 2021.

\bibitem[Rashid et~al.(2020{\natexlab{a}})Rashid, Farquhar, Peng, and Whiteson]{rashid2020weighted}
Tabish Rashid, Gregory Farquhar, Bei Peng, and Shimon Whiteson.
\newblock Weighted qmix: Expanding monotonic value function factorisation for deep multi-agent reinforcement learning.
\newblock \emph{Advances in neural information processing systems}, 33:\penalty0 10199--10210, 2020{\natexlab{a}}.

\bibitem[Rashid et~al.(2020{\natexlab{b}})Rashid, Samvelyan, De~Witt, Farquhar, Foerster, and Whiteson]{rashid2020monotonic}
Tabish Rashid, Mikayel Samvelyan, Christian~Schroeder De~Witt, Gregory Farquhar, Jakob Foerster, and Shimon Whiteson.
\newblock Monotonic value function factorisation for deep multi-agent reinforcement learning.
\newblock \emph{Journal of Machine Learning Research}, 21\penalty0 (178):\penalty0 1--51, 2020{\natexlab{b}}.

\bibitem[Rebello et~al.(2023)Rebello, Tang, Wiens, and Parbhoo]{rebello2023leveraging}
Aaman~Peter Rebello, Shengpu Tang, Jenna Wiens, and Sonali Parbhoo.
\newblock Leveraging factored action spaces for off-policy evaluation.
\newblock In \emph{ICML Workshop on New Frontiers in Learning, Control, and Dynamical Systems}, 2023.

\bibitem[Seyde et~al.(2021)Seyde, Gilitschenski, Schwarting, Stellato, Riedmiller, Wulfmeier, and Rus]{seyde2021bang}
Tim Seyde, Igor Gilitschenski, Wilko Schwarting, Bartolomeo Stellato, Martin Riedmiller, Markus Wulfmeier, and Daniela Rus.
\newblock Is bang-bang control all you need? solving continuous control with {Bernoulli} policies.
\newblock \emph{Advances in Neural Information Processing Systems}, 34:\penalty0 27209--27221, 2021.

\bibitem[Seyde et~al.(2022)Seyde, Werner, Schwarting, Gilitschenski, Riedmiller, Rus, and Wulfmeier]{seyde2022solving}
Tim Seyde, Peter Werner, Wilko Schwarting, Igor Gilitschenski, Martin Riedmiller, Daniela Rus, and Markus Wulfmeier.
\newblock Solving continuous control via {Q-learning}.
\newblock In \emph{The Eleventh International Conference on Learning Representations}, 2022.

\bibitem[Sharma et~al.(2017)Sharma, Suresh, Ramesh, and Ravindran]{sharma2017learning}
Sahil Sharma, Aravind Suresh, Rahul Ramesh, and Balaraman Ravindran.
\newblock Learning to factor policies and action-value functions: Factored action space representations for deep reinforcement learning.
\newblock \emph{arXiv preprint arXiv:1705.07269}, 2017.

\bibitem[Sunehag et~al.(2017)Sunehag, Lever, Gruslys, Czarnecki, Zambaldi, Jaderberg, Lanctot, Sonnerat, Leibo, Tuyls, et~al.]{sunehag2017value}
Peter Sunehag, Guy Lever, Audrunas Gruslys, Wojciech~Marian Czarnecki, Vinicius Zambaldi, Max Jaderberg, Marc Lanctot, Nicolas Sonnerat, Joel~Z Leibo, Karl Tuyls, et~al.
\newblock Value-decomposition networks for cooperative multi-agent learning.
\newblock \emph{arXiv preprint arXiv:1706.05296}, 2017.

\bibitem[Sutton and Barto(2018)]{sutton2018reinforcement}
Richard~S Sutton and Andrew~G Barto.
\newblock Reinforcement learning: An introduction.
\newblock \emph{MIT press}, 2018.

\bibitem[Swazinna et~al.(2021)Swazinna, Udluft, and Runkler]{swazinna2021overcoming}
Phillip Swazinna, Steffen Udluft, and Thomas Runkler.
\newblock Overcoming model bias for robust offline deep reinforcement learning.
\newblock \emph{Engineering Applications of Artificial Intelligence}, 104:\penalty0 104366, 2021.

\bibitem[Tang et~al.(2022)Tang, Makar, Sjoding, Doshi-Velez, and Wiens]{tang2022leveraging}
Shengpu Tang, Maggie Makar, Michael Sjoding, Finale Doshi-Velez, and Jenna Wiens.
\newblock Leveraging factored action spaces for efficient offline reinforcement learning in healthcare.
\newblock \emph{Advances in Neural Information Processing Systems}, 35:\penalty0 34272--34286, 2022.

\bibitem[Tang and Agrawal(2020)]{tang2020discretizing}
Yunhao Tang and Shipra Agrawal.
\newblock Discretizing continuous action space for on-policy optimization.
\newblock In \emph{Proceedings of the aaai conference on artificial intelligence}, volume~34, pages 5981--5988, 2020.

\bibitem[Tavakoli et~al.(2018)Tavakoli, Pardo, and Kormushev]{tavakoli2018action}
Arash Tavakoli, Fabio Pardo, and Petar Kormushev.
\newblock Action branching architectures for deep reinforcement learning.
\newblock In \emph{Proceedings of the aaai conference on artificial intelligence}, volume~32, 2018.

\bibitem[Thrun and Schwartz(1993)]{thrun1993issues}
Sebastian Thrun and Anton Schwartz.
\newblock Issues in using function approximation for reinforcement learning.
\newblock In \emph{Proceedings of the Fourth Connectionist Models Summer School}, volume 255, page 263. Hillsdale, NJ, 1993.

\bibitem[Tunyasuvunakool et~al.(2020)Tunyasuvunakool, Muldal, Doron, Liu, Bohez, Merel, Erez, Lillicrap, Heess, and Tassa]{tunyasuvunakool2020dm_control}
Saran Tunyasuvunakool, Alistair Muldal, Yotam Doron, Siqi Liu, Steven Bohez, Josh Merel, Tom Erez, Timothy Lillicrap, Nicolas Heess, and Yuval Tassa.
\newblock dm\_control: Software and tasks for continuous control.
\newblock \emph{Software Impacts}, 6:\penalty0 100022, 2020.

\bibitem[Van~de Wiele et~al.(2020)Van~de Wiele, Warde-Farley, Mnih, and Mnih]{van2020q}
Tom Van~de Wiele, David Warde-Farley, Andriy Mnih, and Volodymyr Mnih.
\newblock Q-learning in enormous action spaces via amortized approximate maximization.
\newblock \emph{arXiv preprint arXiv:2001.08116}, 2020.

\bibitem[Wu et~al.(2019)Wu, Tucker, and Nachum]{wu2019behavior}
Yifan Wu, George Tucker, and Ofir Nachum.
\newblock Behavior regularized offline reinforcement learning.
\newblock \emph{arXiv preprint arXiv:1911.11361}, 2019.

\bibitem[Yang et~al.(2022)Yang, Bai, Ma, Wang, Zhang, and Han]{yangrorl}
Rui Yang, Chenjia Bai, Xiaoteng Ma, Zhaoran Wang, Chongjie Zhang, and Lei Han.
\newblock {RORL}: Robust offline reinforcement learning via conservative smoothing.
\newblock In \emph{Advances in Neural Information Processing Systems}, 2022.

\bibitem[Yu et~al.(2021{\natexlab{a}})Yu, Liu, Nemati, and Yin]{yu2021reinforcement}
Chao Yu, Jiming Liu, Shamim Nemati, and Guosheng Yin.
\newblock Reinforcement learning in healthcare: A survey.
\newblock \emph{ACM Computing Surveys (CSUR)}, 55\penalty0 (1):\penalty0 1--36, 2021{\natexlab{a}}.

\bibitem[Yu et~al.(2020)Yu, Thomas, Yu, Ermon, Zou, Levine, Finn, and Ma]{yu2020mopo}
Tianhe Yu, Garrett Thomas, Lantao Yu, Stefano Ermon, James~Y Zou, Sergey Levine, Chelsea Finn, and Tengyu Ma.
\newblock {MOPO}: Model-based offline policy optimization.
\newblock \emph{Advances in Neural Information Processing Systems}, 33:\penalty0 14129--14142, 2020.

\bibitem[Yu et~al.(2021{\natexlab{b}})Yu, Kumar, Rafailov, Rajeswaran, Levine, and Finn]{yu2021combo}
Tianhe Yu, Aviral Kumar, Rafael Rafailov, Aravind Rajeswaran, Sergey Levine, and Chelsea Finn.
\newblock {COMBO}: Conservative offline model-based policy optimization.
\newblock \emph{Advances in neural information processing systems}, 34:\penalty0 28954--28967, 2021{\natexlab{b}}.

\bibitem[Zhao et~al.(2018)Zhao, Xia, Zhang, Ding, Yin, and Tang]{zhao2018deep}
Xiangyu Zhao, Long Xia, Liang Zhang, Zhuoye Ding, Dawei Yin, and Jiliang Tang.
\newblock Deep reinforcement learning for page-wise recommendations.
\newblock In \emph{Proceedings of the 12th ACM conference on recommender systems}, pages 95--103, 2018.

\bibitem[Zhou et~al.(2021)Zhou, Bajracharya, and Held]{zhou2021plas}
Wenxuan Zhou, Sujay Bajracharya, and David Held.
\newblock {PLAS}: Latent action space for offline reinforcement learning.
\newblock In \emph{Conference on Robot Learning}, pages 1719--1735. PMLR, 2021.

\end{thebibliography}

\newpage
 \appendix
 \section*{Appendix}

\section{DQN vs DecQN for uniformly distributed errors}\label{sec: uniform_noise}

    To provide support for the ideas presented in Section \ref{Case}, we investigate properties of the target difference under DQN and DecQN based on uniformly distribution noise.
    
    As demonstrated by \cite{thrun1993issues} and \cite{ireland2023revalued}, if $\epsilon(s, \mathbf{a})$ are modelled as independent identically distributed (i.i.d.) uniform random variables $U(-b, b)$, then for $|A|$ actions the expectation and variance of the target difference are, respectively,
    \begin{equation}\label{TarDifExp}
        \mathbb{E}[Z_s] = \gamma b \frac{|A| - 1}{|A| + 1},
    \end{equation}
    \begin{equation}\label{TarDifVar}
        Var(Z_s) = \gamma 4 b^2 \frac{|A|}{(|A| + 1)^2 (|A| + 2)}.
    \end{equation}
    For $|A| > 1$, i.e. more than one action, $\mathbb{E}[Z_s]$ is positive and increasing w.r.t. $|A|$ whereas $Var(Z_s)$ is positive and decreasing.  Hence there is a bias/variance trade-off for overestimation in target Q-values.

    This trade-off becomes apparent when comparing overestimation for DQN and DecQN, where $|A| = \prod_{i=1}^N n_i$ for the former and $|A| = \sum_{i=1}^N n_i$ for the latter.  As shown by \cite{ireland2023revalued}, if $\epsilon^i(s, a_i)$ are also modelled as i.i.d. uniform random variables $U(-b, b)$, then the expected target difference under DecQN is lower than DQN ($\mathbb{E}[ Z_{s'}^{dec}] \le \mathbb{E}[Z_{s'}^{dqn}]$), but the variance is higher ($V[ Z_{s'}^{dec}] \ge V[Z_{s'}^{dqn}]$).  
    
    We now consider this bias/variance trade-off in the context of in-distribution and out-of-distribution actions/sub-actions.  This time, let $U(-b, b)$ and $U(-kb, kb)$ be the distribution of errors for in-distribution and out-of-distribution actions/sub-actions, respectively, where $k > 1$.  When all actions/sub-actions are either in-distribution or out-of-distribution the above conclusions hold, with the additional observation that both bias and variance are lower when all actions are in-distribution compared to all out-of-distribution (as $b < kb$).

    The picture becomes more complex when there is a mixture of in-distribution and out-of-distribution actions/sub-actions.  This stems from the fact there is no longer a closed form for the expectation/variance of the target difference since errors arise from two different distributions.  This is exacerbated by the relative coverage of sub-actions and atomic actions.  In terms of expectation, since this is an increasing function of both $b$ and $|A|$ we can deduce the property $\mathbb{E}[ Z_{s'}^{dec}] \le \mathbb{E}[Z_{s'}^{dqn}]$ holds regardless of the value of $k$ or the number of in-distribution actions/sub-actions.  The variance is more nuanced since this is an increasing function for $b$ but decreasing for $|A|$.  The value of $k$ and number of in-distribution actions/sub-actions will determine whether the property $V[Z_{s'}^{dec}] \ge V[Z_{s'}^{dqn}]$ still holds.

    To illustrate this, we conduct a series of simulations based on different action/sub-action configurations and compare how the expectation and variance of the target difference under DQN and DecQN change based on the coverage of atomic actions.  The details are as follows.
    
    \textbf{DQN simulations:}  Let $|A^{in}|$ and $|A^{out}|$ be the number of in-distribution and out-of-distribution atomic actions for a given state, respectively, such that $|A| = |A^{in}| + |A^{out}|$.  For $|A^{in}| \in \{ 0, 1, ... , |A| \}$ sample $|A^{in}|$ values from a $U(-b, b)$ distribution and $|A| - |A^{in}|$ values from a $U(-kb, kb)$ distribution and store the maximum value for the pooled sample.  Repeat this 10000 times and calculate the mean and variance of these stored maximums.  These are the estimates of $\mathbb{E}[Z_{s}^{dqn}]$ and $V[Z_{s}^{dqn}]$.
    
    \textbf{DecQN simulations}  Let $|A_i^{in}|$ and $|A_i^{out}|$ be the number of in-distribution and out-of-distribution sub-actions in dimension $i$ for a given state, respectively, such that $|A_i| = |A_i^{in}| + |A_i^{out}|$.  For $|A^{in}| \in \{ 0, 1, ... , |A| \}$ sample $|A^{in}|$ atomic action from all available actions without replacement and calculate the number of factorised actions $|A_i^{in}|$ in each sub-action dimension $i$.  In each sub-action dimension $i$ sample $|A_i^{in}|$ values from a $U(-b, b)$ distribution and $|A_i| - |A_i^{in}|$ values from a $U(-kb, kb)$ distribution and store the maximum value for the pooled sample for each dimension $i$.  Calculate and record the average maximum value across all dimensions $N$ (as per the DecQN decomposition).  Repeat this 10000 times and calculate the mean and variance for these stored maximums.  Since sub-action coverage is dependent on which atomic actions are sampled, we repeat this entire procedure 100 times to sample a broad range of atomic actions and calculate the overall mean and variance across these 100 simulations.  These are the estimates of $\mathbb{E}[Z_{s}^{dec}]$ and $V[Z_{s}^{dec}]$.

\begin{figure}[ht]
            \centering
            \includegraphics[width=\linewidth]{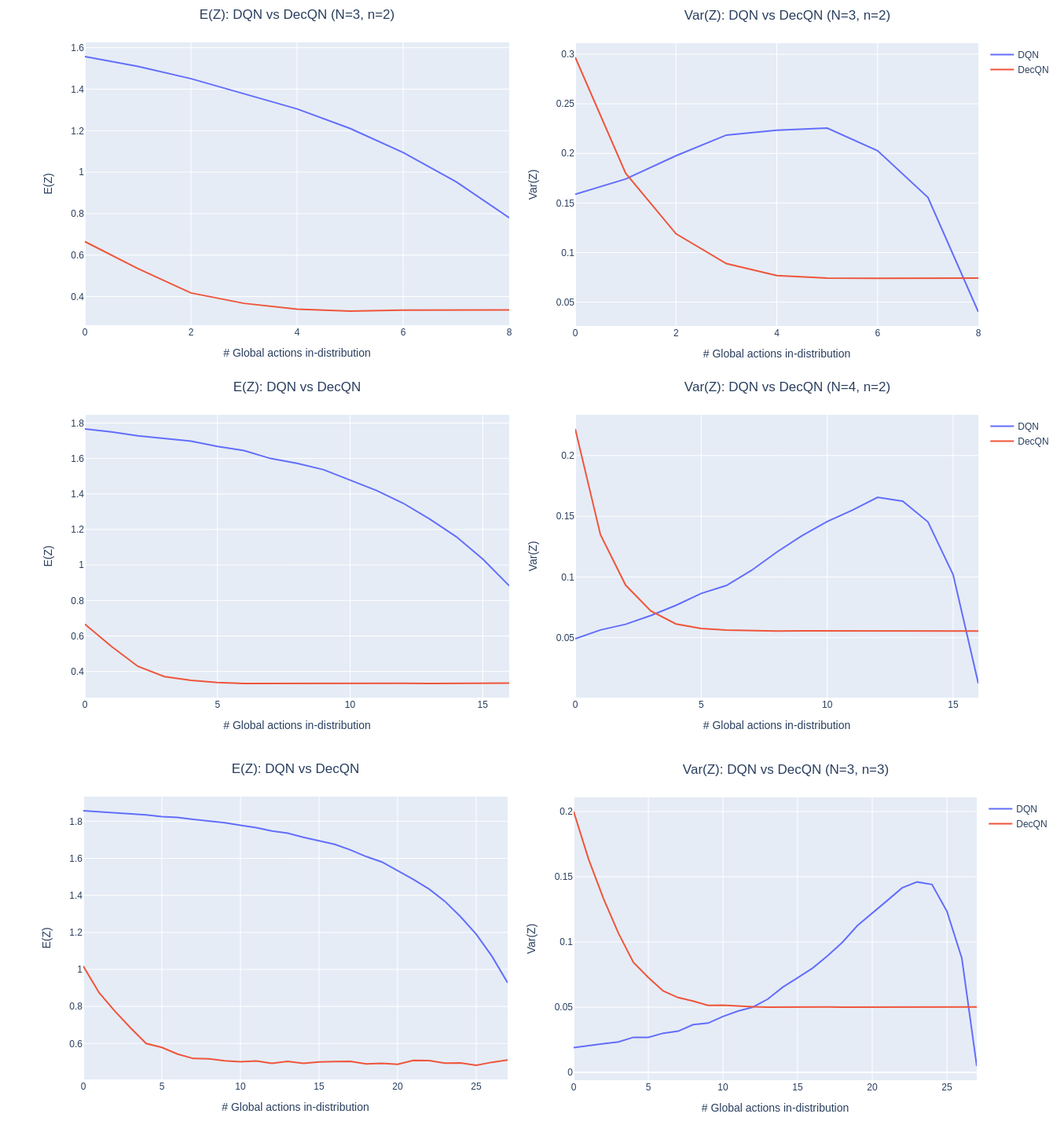}
            \caption{Comparing the expectation and variance of target differences under DQN and DecQN for various configurations and coverages of atomic/sub-actions. \label{fig: target_diff}}
        \end{figure}

    In Figure \ref{fig: target_diff} we summarise these simulations for three configurations: $N=3$ each with two sub-actions, $N=4$ each with two sub-actions and $N=3$ each with three sub-actions.  We set $b=1$ and $k=2$.  For the expectation, we see that in all cases $\mathbb{E}[ Z_{s'}^{dec}] \le \mathbb{E}[Z_{s'}^{dqn}]$ for all values of $|A^{in}|$.  For the variance, we see that in all cases $V[Z_{s'}^{dec}] \ge V[Z_{s'}^{dqn}]$ for values of $|A^{in}|$ close to $0$ and $|A|$ but in between $V[Z_{s'}^{dec}] \le V[Z_{s'}^{dqn}]$.  This aligns with our intuition, namely that the expectation of the target difference under DecQN will always be lower than DQN, but the variance will depend on the relative proportions of in-distribution actions/sub-actions under atomic and factorised representations.  Note that if all atomic actions are in-distribution or out-of-distribution (i.e. $|A^{in}| = 0$ or $|A^{in}| = |A|$) the expectation and variance revert back to Equations \ref{TarDifExp} and \ref{TarDifVar}.

\section{Algorithms - full procedures}\label{appendix:algos}
This section contains the full procedures for algorithms detailed in Section \ref{algos}.

    \begin{algorithm}[ht]
            \caption{DecQN-BCQ}
            \label{alg:DecQN+BCQ}
            \begin{algorithmic}
                \Require Threshold $\tau$, discounter factor $\gamma$, target network update rate $\mu$, number sub-action spaces $N$ and dataset $\mathcal{B}$.
                \State Initialise utility function parameters $\theta = \{\theta_{i}\}^N_{i=1}$, corresponding target parameters $\hat{\theta} = \theta$ and policy parameters $\phi = \{\phi_i\}^N_{i=1}$
                \For {$t=0$ to $T$}
                \State Sample minibatch of transitions $(s, \mathbf{a}, r, s')$ from $\mathcal{B}$
                \State $\phi \leftarrow \argmin_\phi \frac{1}{N} \sum^N_{i=1} -\sum_{s, a_i} \log \pi^i_{\phi_i} (a_i \mid s)$
                \State $\theta \leftarrow \argmin_\theta \sum_{s, \mathbf{a}, r, s'} (Q_\theta(s, \mathbf{a}) - y)^2$
                \State where:
                \State \; $Q_\theta(s, \mathbf{a}) = 1/N \sum^N_{i=1} U^i_{\theta_i}(s, a_i)$,
                \State \; $y = r + \gamma/N \sum^N_{i=1} \max_{a'_i \; : \; \rho^i(a'_i) \geq \tau} U^i_{\hat{\theta_i}} (s', a'_i) $,
                \State \; \; $\rho^i(a'_i) = \pi^i_{\phi_i}(a'_i \mid s') / \max_{\hat{a}'_i} \pi^i_{\phi _i}(\hat{a}'_i \mid s')$
                \State $\hat{\theta} \leftarrow \mu \theta + (1 - \mu)\hat{\theta}$
                \EndFor
            \end{algorithmic}
        \end{algorithm}

    \begin{algorithm}[ht]
    \caption{DecQN-CQL}\label{alg:DecQN+CQL}
    \begin{algorithmic}
    \Require Conservative coefficient $\alpha$, discounter factor $\gamma$, target network update rate $\mu$, number sub-action spaces $N$ and dataset $\mathcal{B}$.
    \State Initialise utility function parameters $\theta = \{\theta_{i}\}^N_{i=1}$ and  corresponding target parameters $\hat{\theta} = \theta$
    \For {$t=0$ to $T$}
       \State Sample minibatch of transitions $(s, \mathbf{a}, r, s')$ from $\mathcal{B}$
       \State $\theta \leftarrow \argmin_\theta \sum_{s, \mathbf{a}, r, s'} (Q_\theta(s, \mathbf{a}) - y)^2 + \alpha \sum_{s, \mathbf{a}} \frac{1}{N} \sum^N_{i=1} \big[ \log \sum_{a_j \in \mathcal{A}_i} \exp (U^i_{\theta_i} (s, a_{j})) - U^i_{\theta_i}(s, a_i) \big]$
       \State where:
       \State \; $Q_\theta(s, \mathbf{a}) = 1/N \sum^N_{i=1} U^i_{\theta_i}(s, a_i)$,
       \State \; $y = r + 1/N \sum^N_{i=1} \max_{a'_i} U^i_{\hat{\theta_i}}(s', a'_i)$    
       \State $\hat{\theta} \leftarrow \mu \theta + (1 - \mu)\hat{\theta}$
    \EndFor
    \end{algorithmic}
    \end{algorithm}

    \begin{algorithm}[ht]
    \caption{DecQN-IQL}\label{alg:DecQN+IQL}
    \begin{algorithmic}
    \Require Expectile $\tau$, discounter factor $\gamma$, target network update rate $\mu$, number sub-action spaces $N$ and dataset $\mathcal{B}$.
    \State Initialise utility function parameters $\theta = \{\theta_{i}\}^N_{i=1}$ and  corresponding target parameters $\hat{\theta} = \theta$.  Initialise state value function parameters $\psi$ and policy parameters $\phi = \{\phi_i\}^N_{i=1}$
    \For {$t=0$ to $T$}
       \State Sample minibatch of transitions $(s, \mathbf{a}, r, s')$ from $\mathcal{B}$
       \State $\phi \leftarrow \argmin_\phi \frac{1}{N} \sum^N_{i=1} -\sum_{s, a_i} \log \pi^i_{\phi_i} (a_i \mid s)$
       \State $\theta \leftarrow \argmin_\theta \sum_{s, \mathbf{a}, r, s'} (Q_\theta(s, \mathbf{a}) - y)^2$
       \State $\psi \leftarrow \argmin_\psi \sum_{s, \mathbf{a}} [ L^\tau_2 (Q_{\hat{\theta}}(s, \mathbf{a}) - V_\psi(s)) ]$
       \State where:
       \State \; $Q_\theta(s, \mathbf{a}) = 1/N \sum^N_{i=1} U^i_{\theta_i}(s, a_i)$,
       \State \; $Q_{\hat{\theta}}(s, \mathbf{a}) = 1/N \sum^N_{i=1} U^i_{\hat{\theta_i}}(s, a_i)$,
       \State \; $y = r + V_\psi(s')$  
       \State $\hat{\theta} \leftarrow \mu \theta + (1 - \mu)\hat{\theta}$
    \EndFor
    \end{algorithmic}
    \end{algorithm}

    \begin{algorithm}[ht]
    \caption{DecQN-OneStep}\label{alg:DecQN+OneStep}
    \begin{algorithmic}
    \Require Discounter factor $\gamma$, target network update rate $\mu$, number sub-action spaces $N$ and dataset $\mathcal{B}$.
    \State Initialise utility function parameters $\theta = \{\theta_{i}\}^N_{i=1}$ and  corresponding target parameters $\hat{\theta} = \theta$.  Initialise policy parameters $\phi = \{\phi_i\}^N_{i=1}$
    \For {$t=0$ to $T$}
       \State Sample minibatch of transitions $(s, \mathbf{a}, r, s')$ from $\mathcal{B}$
       \State $\phi \leftarrow \argmin_\phi \frac{1}{N} \sum^N_{i=1} -\sum_{s, a_i} \log \pi^i_{\phi_i} (a_i \mid s)$
       \State $\theta \leftarrow \argmin_\theta \sum_{s, \mathbf{a}, r, s'} (Q_\theta(s, \mathbf{a}) - y)^2$
       \State where:
       \State \; $Q_\theta(s, \mathbf{a}) = 1/N \sum^N_{i=1} U^i_{\theta_i}(s, a_i)$,
       \State \; $y = r + 1/N \sum^N_{i=1} \sum_{a_i} \pi^i_{\phi_i} (a_i \mid s) U^i_{\hat{\theta_i}}(s', a'_i)$
       \State $\hat{\theta} \leftarrow \mu \theta + (1 - \mu)\hat{\theta}$
    \EndFor
    \end{algorithmic}
    \end{algorithm}

\section{Data collection procedures}
    \label{sec: data collection procedure}

    \subsection{Maze}
    For the Maze environment, we used the default settings as per \citep{chandak2019learning} with the exception that all actions are available from the beginning.  To collect the datasets, we first trained agents for varying numbers of actuators using DecQN, parameterising utility functions as 2-layer MLPs each with 512 nodes that take in a state and output sub-action utility values.  We maintained a fixed exploration parameter $\epsilon = 0.1$ and updated target network parameters using Polyak-averaging with $\mu=0.005$.  We used the Adam optimiser \citep{kingma2014adam} with learning rate $3e-4$ and a batch size of 256.

    Once we have trained the DecQNs we create the benchmark datasets by collecting data using a greedy policy derived from the learned utility values.  Each dataset contains 10k transitions.  For the expert policy we trained the DecQNs until their test performance approached the maximum return possible (approximately 100).  For the medium policy we aimed for a test performance of approximately 1/3rd of the expert.

    \subsection{DeepMind Control Suite}
    To collect the DMC suite datasets we followed the training procedures laid out by \citep{seyde2022solving, ireland2023revalued} to train the Decoupled Q-networks. To expedite the data collection process during training we used a distributed setup using multiple workers to collect data in parallel \citep{horgan2018distributed}. We parameterised the utility functions using a (shared) single ResNet layer followed by layer norm, followed by a linear head for each of the sub-action spaces which predicts sub-action utility values.  Full details regarding network architecture and hyperparameters can be found in Table \ref{tab: collection_hyperparameters}.
    
    Once we have trained the DecQNs we create the benchmark datasets by collecting data using a greedy policy derived from the learned utility values. For the case study in Section \ref{AtomicVsDecQN} each dataset contains 100k transitions and for the main experiments in Section \ref{main_results} each dataset contains 1M transitions.  As each episode is truncated at 1,000 time steps in the DM control suite, this corresponds to collecting 100 episodes for the case study and 1,000 episodes for the main experiments . For the expert policy we trained the DecQNs until their test performance corresponded to the performance given in \citep{seyde2022solving, ireland2023revalued}. For the medium policy we aimed for a test performance of approximately 1/3rd of the reported expert score.
    
    We largely employ the same hyperparameters as the original DecQN study, as detailed in Table \ref{tab: collection_hyperparameters}. Exceptions include the decay of the exploration parameter ($\epsilon$) to a minimum value instead of keeping it constant, and the use of Polyak-averaging for updating the target network parameters, as opposed to a hard reset after every specified number of updates. Finally, we sample from the replay buffer uniformly at random, as opposed to using a priority. We maintain the same hyperparameters across all our experiments.

    For $n=3$ we use DecQN to train networks and collect datasets.  For $n>3$ we use REValueD to train networks and collect datasets due to better scaling to higher numbers of bins \citep{ireland2023revalued}.
    
    \begin{table}[ht]
    \caption{Hyperparameters used in DecQN and REValueD training.\label{tab: collection_hyperparameters}}
     \centering
     \begin{tabular}{cc}
         \toprule
         Parameters & Value \\
         \midrule
         Optimizer & Adam \\
         Learning rate & $1 \times 10^{-4}$ \\
         Replay size & $5 \times 10^{5}$ \\
         n-step returns & 3 \\
         Discount, $\gamma$ & 0.99 \\
         Batch size & 256 \\
         Hidden size & 512 \\
         Gradient clipping & 40 \\
         Target network update parameter, $c$ & 0.005 \\
         Imp. sampling exponent & 0.2 \\
         Priority exponent & 0.6 \\
         Minimum exploration, $\epsilon$ & 0.05 \\
         $\epsilon$ decay rate & 0.99995 \\
         \midrule
         Regularisation loss coefficient $\beta$ & 0.5 \\
         Ensemble size K & 10 \\
         \bottomrule
     \end{tabular}
    \end{table}

\section{Full implementation details}\label{appendix:full_imp}
For both Maze and DMC environments/tasks, utility functions are parameterised by neural networks, comprising a 2-layer MLP with ReLU activation functions and 512 nodes, taking in a normalised state as input and outputting utility values for each sub-action space.  We use the same architecture for policies, with the output layer a softmax across actions within each sub-actions-space.  State value functions mirror this architecture except in the final layer which outputs a single value.  We train networks via stochastic gradient descent using the Adam optimiser \citep{kingma2014adam} with learning rate $3e^{-4}$ and a batch size of 256.  For state and state-action value functions we use the Huber loss as opposed to MSE loss.  We set the discount factor $\gamma=0.99$ and the target network update rate $\mu=0.005$.  We utilise a dual-critic approach, taking the mean across two utility estimates for target Q-values.  For the maze task, agents are trained for 100k gradient updates.  For the DMC tasks, agents are trained for 1M gradient updates.
        
The only hyperparameters we tune are the threshold $\tau$ in BCQ, conservative coefficent $\alpha$ in CQL, expectile $\tau$ and balance coefficient $\lambda$ in IQL and balance coefficient $\lambda$ in OneStep.  We allow these to vary across environment/task, but to better reflect real-world scenarios where the quality of data may be unknown, we forbid variation within environments/tasks.  

Table \ref{tab: hyperparameters} provides hyperparameters for all environments/tasks.  For DecQN-BCQ we searched over $\tau = \{0.025, 0.05, 0.1, 0.25, 0.5, 0.75\}$.  For DecQN-CQL we searched over $\alpha = \{0.25, 0.5, 1, 2\}$.  For DecQN-IQL we searched over $\tau = \{0.5, 0.6, 0.7, 0.8\}$, $\lambda=\{1, 2, 5, 10, 20, 50\}$.  For DecQN-OneStep we searched over $\lambda=\{1, 2, 5, 10, 20, 50\}$.

    \begin{table}[ht]
    \caption{Hyperparameters for experiments in Section \ref{experiments} \label{tab: hyperparameters}}
    \centering
    \begin{tabular}{cccccc}
    \toprule
    Environment/task & Number of bins ($n$) & BCQ $\tau$ & CQL $\alpha$ & IQL $\beta$, $\lambda$ & OneStep $\lambda$ \\
    \midrule
    Maze             & 3              & 0.5        & 1           & 0.5, 20                 & 50                 \\
    Maze             & 5              & 0.5        & 0.25         & 0.5, 20                & 20                 \\
    Maze             & 7              & 0.5        & 0.5          & 0.5, 50                & 20                 \\
    Maze             & 10             & 0.5        & 0.5          & 0.5, 50                & 20                 \\
    Maze             & 12             & 0.5        & 0.5          & 0.5, 50                & 50                 \\
    Maze             & 15             & 0.5        & 0.5          & 0.5, 20                & 20                 \\
    FingerSpin       & 3              & 0.5        & 0.25         & 0.5, 1                 & 1                 \\
    FishSwim         & 3              & 0.25       & 0.25         & 0.5, 1                 & 1                 \\
    CheetahRun       & 3              & 0.05       & 0.25         & 0.5, 1                 & 1                 \\
    QuadrupedWalk    & 3              & 0.25       & 0.25         & 0.5, 2                 & 1                 \\
    HumanoidStand    & 3              & 0.5        & 0.25         & 0.5, 2                 & 2                 \\
    DogTrot          & 3              & 0.25       & 1            & 0.5, 5                 & 5                 \\
    DogTrot          & 10             & 0.25       & 0.25         & 0.5, 5                 & 2                 \\
    DogTrot          & 30             & 0.75       & 1            & 0.5, 2                 & 2                 \\
    DogTrot          & 50             & 0.5        & 0.5          & 0.5, 2                 & 2                 \\
    DogTrot          & 75             & 0.7        & 0.5          & 0.5, 2                 & 2                 \\
    DogTrot          & 100            & 0.5        & 2            & 0.5, 5                 & 5                 \\
     \bottomrule
    \end{tabular}
    \end{table}

    \subsection{Allowing hyperparameter variation within environment/task}
        \label{sec: allowing variation within task}
    As per Section \ref{DiscConc}, in Table \ref{tab: AppExptHyper} we provide examples of performance improvement after permitting hyperparameter variation within the same environment/task.  In general, we see lower quality datasets benefit from smaller hyperparameters (i.e. those that weight more towards RL and less towards BC) and higher quality datasets benefit from larger hyperparameters (i.e. those the weight more towards BC and less towards RL).  This mirrors findings from previous papers outlined in Section \ref{RelatedWork}.

    \begin{table}[ht]
            \centering
            \caption{Individual performance allowing hyperparameters to vary within environment/taskc; ``dog-trot', $n=3$. Figures are mean normalised scores, with 0 and 100 representing random and expert policies, respectively.  Highest score highlighted in bold \label{tab: AppExptHyper}}
        \begin{tabular}{lcccc}
        \toprule
        Environment  -dataset    & \multicolumn{1}{l}{} & \multicolumn{1}{l}{} & \multicolumn{1}{l}{} & \multicolumn{1}{l}{} \\
        \midrule
        DogTrot (BCQ)            & $\tau=0.025$         & $\tau=0.05$          & $\tau=0.1$           & $\tau=0.25$          \\
        -expert                  & 57.8                 & 85                   & 93.6                 & \textbf{94.4}                 \\
        -medium-expert           & 3.9                  & 10.1                 & 42.7                 & \textbf{74.9}                \\
        -medium                  & 39.8                 & 38.8                 & 34.2                 & \textbf{49.1}                 \\
        -random-medium-expert    & 5                    & 5.6                  & \textbf{9}                    & 0.1                  \\
        \midrule
        DogTrot (CQL)            & $\alpha=0.25$        & $\alpha=0.5$         & $\alpha=1$           & $\alpha=2$           \\
        -expert                  & 90.8                 & 95.7                 & 99.5                 & \textbf{100.2}                \\
        -medium-expert           & 76.6                 & 81.7                 & \textbf{84.8}                 & 75.1                 \\
        -medium                  & \textbf{50.6}                 & 48.3                 & 46.5                 & 45.2                 \\
        -random-medium-expert    & 41.2                 & 40.8                 & \textbf{43.4}                 & 38.6                 \\
        \midrule
        DogTrot (IQL $\tau=0.5$) & $\lambda=1$          & $\lambda=2$          & $\lambda=5$          & $\lambda=10$         \\
        -expert                  & 37.9                 & 82.5                 & 98.9                 & \textbf{99.5}                 \\
        -medium-expert           & 33                   & 64                   & 89.3                 & \textbf{98.6}                 \\
        -medium                  & \textbf{58.8}                 & 56.5                 & 52                   & 47.3                 \\
        -random-medium-expert    & 10.6                 & 28.6                 & 44.1                 & \textbf{44.7}                 \\
        \midrule
        DogTrot (OneStep)        & $\lambda=1$          & $\lambda=2$          & $\lambda=5$          & $\lambda=10$         \\
        -expert                  & 53.3                 & 91.7                 & 101.2                & \textbf{102}                  \\
        -medium-expert           & 44.5                 & 79.4                 & 93.9                 & \textbf{96.6}                 \\
        -medium                  & \textbf{59.3}                 & 57.5                 & 50.2                 & 48                   \\
        -random-medium-expert    & 23.8                 & 43.9                 & 44.9                 & \textbf{45.1}                 \\
        \bottomrule
        \end{tabular}
        \end{table}

\section{Number of actions under atomic and factorised representations}
Table \ref{tab:dm_control_suite} summarises the number of actions requiring value estimation under atomic and factorised representations.

\begin{table}[ht]
            \begin{center}
            \caption{Environment details for DeepMind Control Suite.  $|S|$ represents the size of the state space and $N$ the number of sub-action spaces.  $\prod_i n_i$ is the total number of actions under atomic representation and $\sum_i n_i$ under factorised representation when $n_i=3$.}
    	  \label{tab:dm_control_suite}
                \begin{tabular}{ccccc} \toprule
                    Environment & $|S|$ & $N$ & $\prod_i n_i$ & $\sum_i n_i$ \\ \midrule
                    Finger Spin & 9 & 2 & 9 & 6 \\
                    Fish Swim & 24  & 5 & 243 & 15\\
                    Cheetah Run & 17 & 6 & 729 & 18\\
                    Quadruped Walk  & 78 & 12 & $\approx 530k$ & 36\\
                    Humanoid Stand & 67 & 21 & $\approx 10^{10}$ & 63 \\
                    Dog Trot & 223 & 38 & $\approx 10^{18}$ & 114\\ \bottomrule
                \end{tabular}
            \end{center}
        \end{table}

\section{Case study Q-value errors}\label{CaseStudyErrors}
\begin{figure}[ht]
            \centering
            \includegraphics[width=\linewidth]{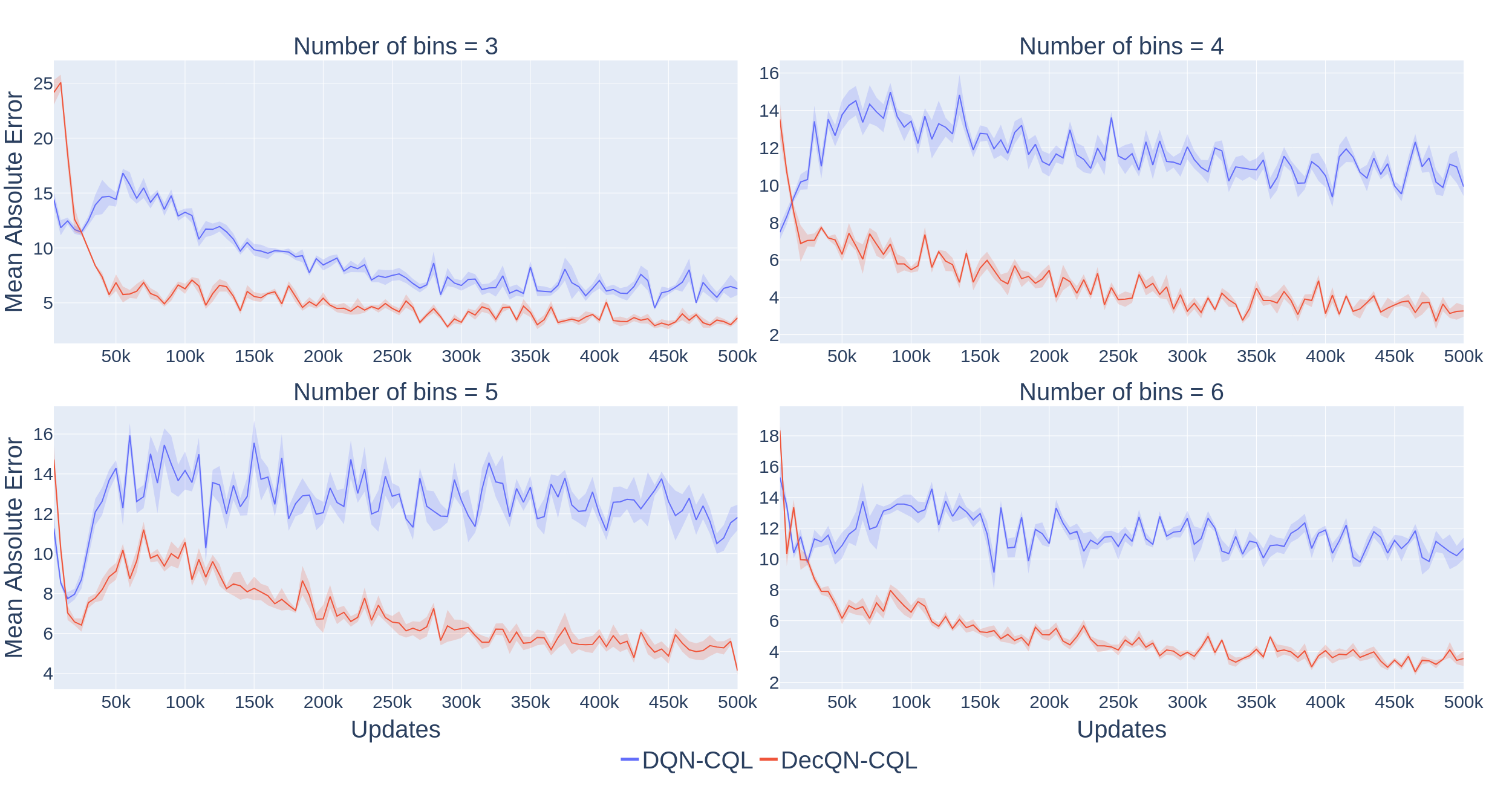}
            \caption{Comparison of estimated errors in Q-values for ``cheetah-run-medium-expert'' dataset for varying numbers of bins.  Errors are lower for DecQN-CQL for all numbers of bins, most notably for $n_i > 3$, mirroring the deviation in performance levels between the two approaches. \label{fig: cql error}}
        \end{figure}

To provide further insights into Q-value errors for the case study in Section \ref{AtomicVsDecQN}, we examine the evolution of Q-value errors during training.  Every 5k gradient updates we obtain a MC estimate of true Q-values using discounted rewards from environmental rollouts.  We then compare these MC estimates with Q-values predicted by both DQN-CQL and DecQN-CQL networks for the respective actions taken. To make better use of rollouts (which can be expensive), we calculate MC estimates and DQN-/DecQN-CQL Q-values for the first 500 states in the trajectory, as using a discount factor of $\gamma=0.99$ discounts rewards by over 99\% for time-steps beyond 500 (and all tasks considered have trajectory length 1000).  In total we perform 10 rollouts, giving 5000 estimates of the error between true and DQN-CQL/DecQN-CQL Q-values.  In Figure \ref{fig: cql error} we plot the mean absolute error over the course of training, with the solid line representing the mean across five random seeds and shaded area the standard error.  For all values of $n_i$ we observe the mean absolute error is less for DecQN-CQL than DQN-CQL, particularly for $n_i > 3$, aligning with each algorithm's respective performance in Figure \ref{fig: FactVsAtomic}. 

\section{Tabulated results}\label{App_Tab_Results}
Tabulated results for Figure \ref{fig: FactVsAtomic} are presented in Table \ref{tab: FactVsAtomic}.  Tabulated results for Figure \ref{fig: FactVsAtomicMaze} are presented in Table \ref{tab: FactVsAtomicMaze}.  Tabulated results for Figure \ref{Fig: MazeResults} are presented in Table \ref{tab:maze_results} Tabulated results for Figure \ref{Fig: MainResults} are presented in Table \ref{tab:main_results}.  Tabulated results for Figure \ref{Fig: LargerBinResults} are presented in Table \ref{tab:main_results_higher}.

\begin{table}[ht]
\begin{center}
\caption{DQN-CQL vs DecQN-CQL - performance and computation comparison for cheetah-run task.  Performance figures are mean normalised scores $\pm$ one standard error, with 0 and 100 representing random and expert policies, respectively.  Computation figures are training time and GPU usage.  Actions figures are total number of actions requiring value estimation based on atomic/factorised representation.\label{tab: FactVsAtomic}}
\begin{tabular}{cccccc}
\toprule
Method     & $n_i$ & Actions & Score           & Training time (mins) & GPU usage (MB) \\
\midrule
DQN-CQL & 3     & 729     & 79.7 $\pm$ 6.0  & 19                   & 266            \\
DQN-CQL & 4     & 4096    & 30.2 $\pm$ 1.7  & 30                   & 412            \\
DQN-CQL & 5     & 15625   & 32.2 $\pm$ 1.4  & 85                   & 958            \\
DQN-CQL & 6     & 46656   & 32.3 $\pm$ 4.0  & 227                  & 2388           \\
DecQN-CQL  & 3     & 18      & 92.7 $\pm$ 3.7  & 19                   & 244            \\
DecQN-CQL  & 4     & 24      & 85.4 $\pm$ 3.8  & 19                   & 246            \\
DecQN-CQL  & 5     & 30      & 79.5 $\pm$ 1.8  & 20                   & 246            \\
DecQN-CQL  & 6     & 36      & 84.3 $\pm$ 2.5  & 20                   & 246            \\
\bottomrule
\end{tabular}
\end{center}
\end{table}

\begin{table}[ht]
\begin{center}
\caption{DQN-CQL vs DecQN-CQL - performance and computation comparison for Maze task with $N=15$ actuators.  Performance figures are mean normalised scores $\pm$ one standard error, with 0 and 100 representing random and expert policies, respectively.  Computation figures are training time and GPU usage.  Actions figures are total number of actions requiring value estimation based on atomic/factorised representation.\label{tab: FactVsAtomicMaze}}
\begin{tabular}{cccccc}
\toprule
Method     & Number of transitions & Actions & Score           & Training time (mins) & GPU usage (MB) \\
\midrule
DQN-CQL &  100    &  32768    & 0.0 $\pm$ 0.0  &     34               &      1728       \\
DQN-CQL &  250    &  32768   & 48.8 $\pm$ 4.9  &      34              &     1728        \\
DQN-CQL &  500    &  32768  & 68.5 $\pm$ 2.7  &      34             &      1728       \\
DQN-CQL &  1000    &  32768  & 73.7 $\pm$ 2.7  &      34             &    1728        \\
DQN-CQL &  2500    & 32768   & 90.7 $\pm$ 2.5  &      34             &    1728        \\
DQN-CQL &  5000    & 32768   & 96.0 $\pm$ 1.4  &     34              &    1728        \\
DecQN-CQL  & 100     & 30      & 31.9 $\pm$ 10.6  &   4                 &    246         \\
DecQN-CQL  & 250     & 30      & 69.2 $\pm$ 3.2  &     4               &   246          \\
DecQN-CQL  & 500     &  30     & 84.9  $\pm$ 2.3  &    4                &    246         \\
DecQN-CQL  & 1000     & 30      & 84.8 $\pm$ 1.5  &    4                &   246          \\
DecQN-CQL  & 2500     & 30      & 96.4 $\pm$ 1.7  &   4                 &   246          \\
DecQN-CQL  & 5000     & 30      & 99.7 $\pm$ 0.2  &   4                 &   246          \\
\bottomrule
\end{tabular}
\end{center}
\end{table}

\begin{table}[ht]
\centering
    \caption{Individual performance comparison for Maze for varying numbers of actuators. Figures are mean normalised scores $\pm$ one standard error, with 0 and 100 representing random and expert policies, respectively.  At the bottom of the table we also provide totals, split by data quality and the entire set.  We see that DecQN-BCQ is the least performant of the offline methods and that DecQN-CQL/IQL/OneStep perform similarly overall and on expert, medium-exert and and medium datasets and DecQN-CQL the best on random-medium-expert datasets. \label{tab:maze_results}}
\begin{adjustbox}{width=\linewidth}
\begin{tabular}{lcccccc}
\toprule
Environment -dataset  & BC                & DecQN            & DecQN-BCQ          & DecQN-CQL         & DecQN-IQL         & DecQN-OneStep     \\
\midrule
Maze ($Actuators=3$)  &                   &                  &                    &                   &                   &                   \\
-expert               & 100.1 $\pm$ 0     & -5.6 $\pm$ 0     & 100.1 $\pm$ 0      & 100.2 $\pm$ 0     & 100.1 $\pm$ 0     & 100.1 $\pm$ 0     \\
-medium-expert        & 98.7 $\pm$ 1      & -5.6 $\pm$ 0     & 99.6 $\pm$ 0.6     & 100.1 $\pm$ 0     & 100.1 $\pm$ 0     & 99.9 $\pm$ 0      \\
-medium               & 55.3 $\pm$ 2.5    & -5.6 $\pm$ 0     & 54.7 $\pm$ 1.7     & 62.8 $\pm$ 5.3    & 54.6 $\pm$ 7      & 57.1 $\pm$ 6.5    \\
-random-medium-expert & 67.4 $\pm$ 11.9   & 15.5 $\pm$ 46.1  & 34 $\pm$ 5.1       & 95.3 $\pm$ 3.9    & 92.2 $\pm$ 2.9    & 95.6 $\pm$ 1.9    \\
\midrule
Maze ($Actuators=5$)  &                   &                  &                    &                   &                   &                   \\
-expert               & 100.1 $\pm$ 0     & -1.7 $\pm$ 0     & 100.1 $\pm$ 0      & 100.1 $\pm$ 0     & 100.1 $\pm$ 0     & 100.1 $\pm$ 0     \\
-medium-expert        & 99.6 $\pm$ 0.6    & -1.7 $\pm$ 0     & 99.4 $\pm$ 0.9     & 100.1 $\pm$ 0     & 99 $\pm$ 2.3      & 100 $\pm$ 0.1     \\
-medium               & 32.3 $\pm$ 4.3    & -1.7 $\pm$ 0     & 32.7 $\pm$ 2.8     & 44.8 $\pm$ 8.4    & 44.8 $\pm$ 6.4    & 44.5 $\pm$ 3.6    \\
-random-medium-expert & 62.5 $\pm$ 12.1   & -1.7 $\pm$ 0     & 21.4 $\pm$ 18      & 99.3 $\pm$ 1.4    & 99.4 $\pm$ 0.5    & 99.7 $\pm$ 0.4    \\
\midrule
Maze ($Actuators=7$)  &                   &                  &                    &                   &                   &                   \\
-expert               & 100.1 $\pm$ 0     & -0.6 $\pm$ 0     & 100.1 $\pm$ 0      & 100.1 $\pm$ 0     & 100.1 $\pm$ 0     & 100.1 $\pm$ 0     \\
-medium-expert        & 99.6 $\pm$ 0.6    & -0.6 $\pm$ 0     & 98.4 $\pm$ 2.1     & 100.1 $\pm$ 0     & 100.1 $\pm$ 0     & 100.1 $\pm$ 0     \\
-medium               & 68.1 $\pm$ 3      & -0.6 $\pm$ 0     & 70.9 $\pm$ 4       & 76.8 $\pm$ 4.2    & 74 $\pm$ 4        & 73 $\pm$ 4.5      \\
-random-medium-expert & 83.3 $\pm$ 8.2    & -0.6 $\pm$ 0     & 17 $\pm$ 12.5      & 99.2 $\pm$ 1.1    & 98.1 $\pm$ 1.5    & 99.2 $\pm$ 0.6    \\
\midrule
Maze ($Actuators=10$) &                   &                  &                    &                   &                   &                   \\
-expert               & 100.1 $\pm$ 0     & 0 $\pm$ 0        & 100.1 $\pm$ 0      & 100.1 $\pm$ 0     & 100.1 $\pm$ 0     & 100.1 $\pm$ 0     \\
-medium-expert        & 98.6 $\pm$ 1.2    & 0 $\pm$ 0        & 99.3 $\pm$ 0.8     & 100.1 $\pm$ 0     & 100 $\pm$ 0       & 100 $\pm$ 0       \\
-medium               & 40.1 $\pm$ 5.8    & 0 $\pm$ 0        & 40 $\pm$ 6.4       & 53.4 $\pm$ 5.8    & 43.8 $\pm$ 5      & 42.2 $\pm$ 6.5    \\
-random-medium-expert & 84.5 $\pm$ 7      & 0 $\pm$ 0        & 50.9 $\pm$ 4.7     & 99.9 $\pm$ 0.5    & 97.8 $\pm$ 3      & 99.4 $\pm$ 0.7    \\
\midrule
Maze ($Actuators=12$) &                   & $\pm$            & $\pm$              & $\pm$             & $\pm$             & $\pm$             \\
-expert               & 100.1 $\pm$ 0     & 0 $\pm$ 0        & 100.1 $\pm$ 0      & 100.1 $\pm$ 0     & 100.1 $\pm$ 0     & 100.1 $\pm$ 0     \\
-medium-expert        & 97.8 $\pm$ 1.6    & 0 $\pm$ 0        & 98.2 $\pm$ 1.5     & 100.1 $\pm$ 0     & 99 $\pm$ 0.7      & 99.8 $\pm$ 0.5    \\
-medium               & 20.7 $\pm$ 3.2    & 0 $\pm$ 0        & 25.3 $\pm$ 6       & 38.1 $\pm$ 4.8    & 26.4 $\pm$ 4.2    & 30.1 $\pm$ 5.8    \\
-random-medium-expert & 79.7 $\pm$ 6.7    & 0 $\pm$ 0        & 65.3 $\pm$ 29.3    & 98.9 $\pm$ 1.1    & 95.7 $\pm$ 3.9    & 98.1 $\pm$ 1.9    \\
\midrule
Maze ($Actuators=15$) &                   &                  &                    &                   &                   &                   \\
-expert               & 100.1 $\pm$ 0     & 0 $\pm$ 0        & 100.1 $\pm$ 0      & 100.1 $\pm$ 0     & 100.1 $\pm$ 0     & 100.1 $\pm$ 0     \\
-medium-expert        & 100 $\pm$ 0       & 0 $\pm$ 0        & 99.7 $\pm$ 0.6     & 100.1 $\pm$ 0     & 100.1 $\pm$ 0     & 100.1 $\pm$ 0     \\
-medium               & 53.2 $\pm$ 6.2    & 0 $\pm$ 0        & 50.8 $\pm$ 4.6     & 69.1 $\pm$ 3.4    & 91.3 $\pm$ 6.8    & 94.9 $\pm$ 2      \\
-random-medium-expert & 93.1 $\pm$ 3.2    & 0 $\pm$ 0        & 12.1 $\pm$ 17.9    & 99.3 $\pm$ 0.9    & 96.7 $\pm$ 4.9    & 85.1 $\pm$ 12.5   \\
\midrule
Sum                   &                   &                  &                    &                   &                   &                   \\
-expert               & 600.6 $\pm$ 0     & -7.9 $\pm$ 0     & 600.6 $\pm$ 0      & 600.7 $\pm$ 0     & 600.6 $\pm$ 0     & 600.6 $\pm$ 0     \\
-medium-expert        & 594.3 $\pm$ 5     & -7.9 $\pm$ 0     & 594.6 $\pm$ 6.5    & 600.6 $\pm$ 0     & 598.3 $\pm$ 3     & 599.9 $\pm$ 0.6   \\
-medium               & 269.7 $\pm$ 25    & -7.9 $\pm$ 0     & 274.4 $\pm$ 25.5   & 345 $\pm$ 31.9    & 334.9 $\pm$ 33.4  & 341.8 $\pm$ 28.9  \\
-random-medium-expert & 470.5 $\pm$ 49.1  & 13.2 $\pm$ 46.1  & 200.7 $\pm$ 87.5   & 591.9 $\pm$ 8.9   & 579.9 $\pm$ 16.7  & 577.1 $\pm$ 18    \\
-all                  & 1935.1 $\pm$ 79.1 & -10.5 $\pm$ 46.1 & 1670.3 $\pm$ 119.5 & 2138.2 $\pm$ 40.8 & 2113.7 $\pm$ 53.1 & 2119.4 $\pm$ 47.5 \\
\bottomrule
\end{tabular}
\end{adjustbox}
\end{table}

\begin{table}[ht]
    \centering
    \caption{Individual performance comparison for DMC $n=3$. Figures are mean normalised scores $\pm$ one standard error, with 0 and 100 representing random and expert policies, respectively.  At the bottom of the table we also provide totals, split by data quality and the entire set.  We see that DecQN-BCQ is the least performant of the offline methods and that DecQN-CQL/IQL/OneStep perform similarly overall and on expert, medium-exert and and medium datasets and DecQN-CQL the best on random-medium-expert datasets. \label{tab:main_results}}
\begin{adjustbox}{width=\linewidth}
\begin{tabular}{lcccccc}
\toprule
Environment  -dataset & BC                 & DecQN           & DecQN-BCQ         & DecQN-CQL         & DecQN-IQL        & DecQN-OneStep     \\
\midrule
FingerSpin            &                    &                 &                   &                   &                  &                   \\
-expert               & 99.5 $\pm$ 0.4     & -0.2 $\pm$ 0.1  & 100.5 $\pm$ 0.8   & 107.1 $\pm$ 0.3   & 102.9 $\pm$ 0.2  & 102.5 $\pm$ 0.4   \\
-medium-expert        & 77.9 $\pm$ 6.5     & -0.5 $\pm$ 0    & 86.5 $\pm$ 9      & 106.8 $\pm$ 0.2   & 102.8 $\pm$ 0.3  & 102.7 $\pm$ 0.2   \\
-medium               & 38.3 $\pm$ 1       & -0.5 $\pm$ 0    & 40.7 $\pm$ 0.4    & 49.4 $\pm$ 0.3    & 44 $\pm$ 0.8     & 45.2 $\pm$ 0.6    \\
-random-medium-expert & 8.2 $\pm$ 2.3      & 56.2 $\pm$ 2.9  & 62.7 $\pm$ 9.5    & 100 $\pm$ 0.5     & 77.1 $\pm$ 5.1   & 78.5 $\pm$ 3.5    \\
\midrule
FishSwim              &                    &                 &                   &                   &                  &                   \\
-expert               & 82.9 $\pm$ 12.2    & -1.8 $\pm$ 1.9  & 112 $\pm$ 2.7     & 120.8 $\pm$ 14.7  & 123.8 $\pm$ 11.7 & 105.3 $\pm$ 8.2   \\
-medium-expert        & 40.6 $\pm$ 6.9     & 0.7 $\pm$ 6.5   & 97.5 $\pm$ 4.3    & 127.2 $\pm$ 11.1  & 91.1 $\pm$ 13.3  & 112.2 $\pm$ 8.5   \\
-medium               & 42.8 $\pm$ 9.6     & -4 $\pm$ 1.5    & 56.1 $\pm$ 9.8    & 71.5 $\pm$ 6.3    & 63.4 $\pm$ 7.2   & 76.7 $\pm$ 6      \\
-random-medium-expert & 23.6 $\pm$ 9.2     & -0.4 $\pm$ 3.3  & 17.8 $\pm$ 4.8    & 52.1 $\pm$ 9      & 37.1 $\pm$ 9     & 59.9 $\pm$ 11     \\
\midrule
CheetahRun            &                    &                 &                   &                   &                  &                   \\
-expert               & 99.9 $\pm$ 1.7     & 2.7 $\pm$ 2.1   & 105.5 $\pm$ 0.4   & 105.6 $\pm$ 0.9   & 104.6 $\pm$ 1.1  & 106.3 $\pm$ 0.3   \\
-medium-expert        & 61.6 $\pm$ 12.5    & 0.9 $\pm$ 0.8   & 104.2 $\pm$ 1.6   & 103.2 $\pm$ 0.7   & 102.5 $\pm$ 1.2  & 104.8 $\pm$ 0.7   \\
-medium               & 40.4 $\pm$ 0.4     & 0 $\pm$ 0.7     & 47.1 $\pm$ 0.4    & 48.3 $\pm$ 0.3    & 47.7 $\pm$ 0.5   & 47.9 $\pm$ 0.3    \\
-random-medium-expert & 41.5 $\pm$ 0.6     & 0.5 $\pm$ 1.2   & 22.2 $\pm$ 4.4    & 79.6 $\pm$ 5.6    & 61.4 $\pm$ 3.3   & 53.2 $\pm$ 1.9    \\
\midrule
QuadrupedWalk         &                    &                 &                   &                   &                  &                   \\
-expert               & 97.7 $\pm$ 3.2     & 3.3 $\pm$ 4.6   & 114.9 $\pm$ 1.2   & 118.2 $\pm$ 1.4   & 122.3 $\pm$ 1.1  & 115.3 $\pm$ 2.4   \\
-medium-expert        & 63.4 $\pm$ 11      & 8.5 $\pm$ 1.5   & 110.4 $\pm$ 1.9   & 115.4 $\pm$ 4     & 121.2 $\pm$ 1    & 109.4 $\pm$ 2.1   \\
-medium               & 39.2 $\pm$ 8.5     & 11.3 $\pm$ 6.3  & 47.6 $\pm$ 8.4    & 48.6 $\pm$ 7.9    & 46.3 $\pm$ 5.7   & 46.8 $\pm$ 9.2    \\
-random-medium-expert & 28 $\pm$ 6.4       & -5.2 $\pm$ 4.7  & -12.3 $\pm$ 1.5   & 76.7 $\pm$ 5      & 65.8 $\pm$ 4.4   & 69.4 $\pm$ 5.7    \\
\midrule
HumanoidStand         &                    &                 &                   &                   &                  &                   \\
-expert               & 102.2 $\pm$ 1.3    & -0.1 $\pm$ 0    & 103.6 $\pm$ 1.7   & 109 $\pm$ 1.9     & 116.6 $\pm$ 0.6  & 117.2 $\pm$ 0.8   \\
-medium-expert        & 63.1 $\pm$ 3.9     & 0.1 $\pm$ 0     & 92 $\pm$ 2.7      & 104.7 $\pm$ 1.9   & 113.3 $\pm$ 0.4  & 116.7 $\pm$ 0.7   \\
-medium               & 44.4 $\pm$ 0.5     & 0 $\pm$ 0       & 47.2 $\pm$ 1.5    & 51.4 $\pm$ 0.3    & 53.8 $\pm$ 0.4   & 53.8 $\pm$ 0.4    \\
-random-medium-expert & 34.4 $\pm$ 2.6     & 0 $\pm$ 0.1     & 17.7 $\pm$ 5.9    & 42.7 $\pm$ 0.9    & 46 $\pm$ 1       & 47.3 $\pm$ 0.7    \\
\midrule
DogTrot               &                    &                 &                   &                   &                  &                   \\
-expert               & 98 $\pm$ 0.7       & 0.1 $\pm$ 0.1   & 94.4 $\pm$ 0.5    & 99.5 $\pm$ 0.7    & 98.9 $\pm$ 2.2   & 101.2 $\pm$ 1.6   \\
-medium-expert        & 62 $\pm$ 3.7       & 0 $\pm$ 0.3     & 74.9 $\pm$ 3.5    & 84.8 $\pm$ 3.7    & 89.3 $\pm$ 1.4   & 93.9 $\pm$ 1.4    \\
-medium               & 43.8 $\pm$ 0.5     & 0.1 $\pm$ 0.1   & 49.1 $\pm$ 0.6    & 46.5 $\pm$ 0.5    & 52 $\pm$ 0.3     & 50.2 $\pm$ 0.2    \\
-random-medium-expert & 37.2 $\pm$ 3.6     & 0.1 $\pm$ 0.2   & 0.1 $\pm$ 0.1     & 43.4 $\pm$ 0.5    & 44.1 $\pm$ 1.2   & 44.9 $\pm$ 0.7    \\
\midrule
Sum                   &                    &                 &                   &                   &                  &                   \\
-expert               & 580.2 $\pm$ 19.5   & 4 $\pm$ 8.8     & 630.9 $\pm$ 13.2  & 660.2 $\pm$ 19.9  & 669.1 $\pm$ 16.9 & 647.8 $\pm$ 13.7  \\
-medium-expert        & 368.6 $\pm$ 44.5   & 9.7 $\pm$ 9.1   & 565.5 $\pm$ 36.9  & 642.1 $\pm$ 21.6  & 620.2 $\pm$ 17.6 & 639.7 $\pm$ 13.6  \\
-medium               & 248.9 $\pm$ 20.5   & 6.9 $\pm$ 8.6   & 287.8 $\pm$ 44.8  & 315.7 $\pm$ 15.6  & 307.2 $\pm$ 14.9 & 320.6 $\pm$ 16.7  \\
-random-medium-expert & 172.9 $\pm$ 24.7   & 51.2 $\pm$ 12.4 & 108.2 $\pm$ 39.4  & 394.5 $\pm$ 21.5  & 331.5 $\pm$ 24   & 353.2 $\pm$ 23.5  \\
-all                  & 1370.6 $\pm$ 109.2 & 71.8 $\pm$ 38.9 & 1592.4 $\pm$ 134.3& 2012.5 $\pm$ 78.6 & 1928 $\pm$ 73.4  & 1961.3 $\pm$ 67.5 \\
\bottomrule
\end{tabular}
\end{adjustbox}
\end{table}

\begin{table}[ht]
    \centering
    \caption{Individual performance comparison for dog-trot $n=\{3, 10, 30, 50, 75, 100\}$. Figures are mean normalised scores $\pm$ one standard error, with 0 and 100 representing random and expert policies, respectively. At the bottom of the table we also provide totals, split by data quality and the entire set.  We see across all datasets that DecQN-IQL/OneStep perform best, followed by DecQN-CQL and DecQN-BCQ. \label{tab:main_results_higher}}
\begin{adjustbox}{width=\linewidth}
\begin{tabular}{lcccccc}
\toprule
Environment -dataset  & BC                & DecQN         & DecQN-BCQ         & DecQN-CQL         & DecQN-IQL         & DecQN-OneStep     \\
\midrule
DogTrot ($n=3$)       &                   &               &                   &                   &                   &                   \\
-expert               & 98 $\pm$ 0.7      & 0.1 $\pm$ 0.1 & 94.4 $\pm$ 0.5    & 99.5 $\pm$ 0.7    & 98.9 $\pm$ 2.2    & 101.2 $\pm$ 1.6   \\
-medium-expert        & 62 $\pm$ 3.7      & 0 $\pm$ 0.3   & 74.9 $\pm$ 3.5    & 84.8 $\pm$ 3.7    & 89.3 $\pm$ 1.4    & 93.9 $\pm$ 1.4    \\
-medium               & 43.8 $\pm$ 0.5    & 0.1 $\pm$ 0.1 & 49.1 $\pm$ 0.6    & 46.5 $\pm$ 0.5    & 52 $\pm$ 0.3      & 50.2 $\pm$ 0.2    \\
-random-medium-expert & 37.2 $\pm$ 3.6    & 0.1 $\pm$ 0.2 & 0.1 $\pm$ 0.1     & 43.4 $\pm$ 0.5    & 44.1 $\pm$ 1.2    & 44.9 $\pm$ 0.7    \\
\midrule
DogTrot ($n=10$)      &                   &               &                   &                   &                   &                   \\
-expert               & 97.3 $\pm$ 1.7    & -0.3 $\pm$ 0  & 96.8 $\pm$ 1.8    & 99.2 $\pm$ 1.3    & 106.5 $\pm$ 1.1   & 113.4 $\pm$ 0.9   \\
-medium-expert        & 58.1 $\pm$ 5.5    & 0.5 $\pm$ 0.1 & 82.3 $\pm$ 4.6    & 83.4 $\pm$ 2.1    & 105.1 $\pm$ 2.1   & 109.8 $\pm$ 1.2   \\
-medium               & 34.7 $\pm$ 0.4    & 0.5 $\pm$ 0.2 & 36.9 $\pm$ 0.3    & 39.1 $\pm$ 0.6    & 41 $\pm$ 0.8      & 47.2 $\pm$ 0.4    \\
-random-medium-expert & 33.6 $\pm$ 4.4    & 0.1 $\pm$ 0.1 & 6.1 $\pm$ 2.8     & 33.4 $\pm$ 1.5    & 52.1 $\pm$ 2.3    & 45.4 $\pm$ 4.1    \\
\midrule
DogTrot ($n=30$)      &                   &               &                   &                   &                   &                   \\
-expert               & 99.7 $\pm$ 0.4    & -0.3 $\pm$ 0  & 100.3 $\pm$ 2.3   & 99.8 $\pm$ 0.8    & 102.8 $\pm$ 0.4   & 106.4 $\pm$ 0.8   \\
-medium-expert        & 70.3 $\pm$ 4.9    & 0.6 $\pm$ 0.2 & 74.1 $\pm$ 8      & 90.7 $\pm$ 3.7    & 98.3 $\pm$ 1.6    & 100.7 $\pm$ 3     \\
-medium               & 33.1 $\pm$ 0.2    & 0.7 $\pm$ 0.2 & 33.6 $\pm$ 0.4    & 33.4 $\pm$ 0.4    & 39.9 $\pm$ 0.1    & 40.5 $\pm$ 0.5    \\
-random-medium-expert & 22.4 $\pm$ 0.9    & 0.2 $\pm$ 0.1 & 5.2 $\pm$ 2.2     & 22.9 $\pm$ 2.5    & 31.6 $\pm$ 1.3    & 30.4 $\pm$ 1.8    \\
\midrule
DogTrot ($n=50$)      &                   &               &                   &                   &                   &                   \\
-expert               & 98.2 $\pm$ 0.5    & 0.5 $\pm$ 0.1 & 98.6 $\pm$ 0.6    & 97.7 $\pm$ 0.6    & 99.2 $\pm$ 0.9    & 100.4 $\pm$ 1.4   \\
-medium-expert        & 67.7 $\pm$ 5      & -0.3 $\pm$ 0  & 68.7 $\pm$ 4.2    & 88.9 $\pm$ 2.1    & 95 $\pm$ 2.1      & 97.1 $\pm$ 1.4    \\
-medium               & 34.9 $\pm$ 0.4    & -0.3 $\pm$ 0  & 36.5 $\pm$ 0.7    & 36.7 $\pm$ 0.2    & 45.1 $\pm$ 0.6    & 43.1 $\pm$ 0.7    \\
-random-medium-expert & 21 $\pm$ 2        & 0.7 $\pm$ 0.1 & 9.3 $\pm$ 1.5     & 27.6 $\pm$ 1.3    & 27.9 $\pm$ 2.4    & 36.4 $\pm$ 1.8    \\
\midrule
DogTrot ($n=75$)      &                   &               &                   &                   &                   &                   \\
-expert               & 98.4 $\pm$ 0.8    & 0.6 $\pm$ 0.2 & 99.6 $\pm$ 1.1    & 98 $\pm$ 1        & 101.4 $\pm$ 1.4   & 102.9 $\pm$ 0.9   \\
-medium-expert        & 75.3 $\pm$ 2.6    & 0 $\pm$ 0     & 74.9 $\pm$ 16.7   & 85 $\pm$ 2        & 89.9 $\pm$ 0.7    & 97.9 $\pm$ 2.5    \\
-medium               & 41.4 $\pm$ 0.2    & 0.5 $\pm$ 0.1 & 41.3 $\pm$ 0.8    & 42.5 $\pm$ 0.1    & 48.5 $\pm$ 0.4    & 47.8 $\pm$ 0.5    \\
-random-medium-expert & 27.7 $\pm$ 0.7    & 0.7 $\pm$ 0.2 & 0 $\pm$ 0.1       & 33.4 $\pm$ 2      & 45.3 $\pm$ 2.9    & 43.6 $\pm$ 2.6    \\
\midrule
DogTrot ($n=100$)     &                   &               &                   &                   &                   &                   \\
-expert               & 96.4 $\pm$ 1.1    & 0.1 $\pm$ 0   & 95 $\pm$ 1.8      & 100.4 $\pm$ 0.7   & 103.2 $\pm$ 0.4   & 105.5 $\pm$ 0.8   \\
-medium-expert        & 70.9 $\pm$ 6.5    & 0.4 $\pm$ 0.1 & 74.6 $\pm$ 5.8    & 74.4 $\pm$ 5.1    & 93.9 $\pm$ 1      & 101.5 $\pm$ 2.5   \\
-medium               & 34.7 $\pm$ 0.8    & 0.1 $\pm$ 0.1 & 34.5 $\pm$ 0.5    & 34.8 $\pm$ 0.7    & 40.9 $\pm$ 0.2    & 41.5 $\pm$ 0.5    \\
-random-medium-expert & 18.9 $\pm$ 2.7    & 0.3 $\pm$ 0.1 & 0 $\pm$ 0         & 12.7 $\pm$ 2.4    & 24 $\pm$ 2.8      & 26.2 $\pm$ 1      \\
\midrule
Sum                   &                   &               &                   &                   &                   &                   \\
-expert               & 588 $\pm$ 5.2     & 0.7 $\pm$ 0.4 & 584.7 $\pm$ 14    & 594.6 $\pm$ 5.1   & 612 $\pm$ 6.4     & 629.8 $\pm$ 6.4   \\
-medium-expert        & 404.3 $\pm$ 28.2  & 1.2 $\pm$ 0.7 & 449.5 $\pm$ 64.9  & 507.2 $\pm$ 18.7  & 571.5 $\pm$ 8.9   & 600.9 $\pm$ 12    \\
-medium               & 222.6 $\pm$ 2.5   & 1.6 $\pm$ 0.7 & 231.9 $\pm$ 5.8   & 233 $\pm$ 2.5     & 267.4 $\pm$ 2.4   & 270.3 $\pm$ 2.8   \\
-random-medium-expert & 160.8 $\pm$ 14.3  & 2.1 $\pm$ 0.8 & 30.2 $\pm$ 17.6   & 173.4 $\pm$ 10.2  & 225 $\pm$ 12.9    & 226.9 $\pm$ 12    \\
-all                  & 1375.7 $\pm$ 50.2 & 5.6 $\pm$ 2.6 & 1296.3 $\pm$ 102.3& 1508.2 $\pm$ 36.5 & 1675.9 $\pm$ 30.6 & 1727.9 $\pm$ 33.2 \\
\bottomrule
\end{tabular}
\end{adjustbox}
\end{table}

\subsubsection*{Number of seeds}
To check five seeds is sufficient for evaluation purposes we run one task for ten seeds and compare performance in Tables \ref{tab:5seeds} and \ref{tab:10seeds}, respectively.  We see consistent results across different seed counts.  We expect similar consistency across all tasks as the same underlying methodology is applied throughout our experiments.
\begin{table}
    \centering
    \caption{Performance across 5 seeds.\label{tab:5seeds}}
    \begin{tabular}{lcccc}
    \toprule
        Environment -dataset & DecQN-BCQ & DecQN-CQL & DecQN-IQL & DecQN-OneStep \\
    \midrule
        HumanoidStand  &  &  &   &  \\
        -expert & 103.6 $\pm$ 1.7 & 109 $\pm$ 1.9  & 116.6 $\pm$ 0.6  & 117.2 $\pm$ 0.8 \\
        -medium-expert & 92 $\pm$ 2.7  & 104.7 $\pm$ 1.9 & 113.3 $\pm$ 0.4 & 116.7 $\pm$ 0.7\\
        -medium & 47.2 $\pm$ 1.5 & 51.4 $\pm$ 0.3 & 53.8 $\pm$ 0.4 & 53.8 $\pm$ 0.4\\
        -random-medium-expert & 17.7 $\pm$ 5.9 & 42.7 $\pm$ 0.9 & 46 $\pm$ 1 & 47.3 $\pm$ 0.7\\
    \bottomrule
    \end{tabular}
\end{table}

\begin{table}
    \centering
    \caption{Performance across 10 seeds\label{tab:10seeds}}
    \begin{tabular}{lcccc}
    \toprule
        Environment -dataset & DecQN-BCQ & DecQN-CQL & DecQN-IQL & DecQN-OneStep \\
    \midrule
        HumanoidStand  &  &  &   &  \\
        -expert & 103.6 $\pm$ 1.3 & 109.6 $\pm$ 1.5 & 117 $\pm$ 0.6 & 118.1 $\pm$ 0.7 \\
        -medium-expert & 90.3 $\pm$ 3.3  & 103.4 $\pm$ 1.8 & 112.9 $\pm$ 1.1 & 116.2 $\pm$ 0.6\\
        -medium & 46.7 $\pm$ 0.6 & 51.9 $\pm$ 0.4 & 53.7 $\pm$ 0.4 & 53.8 $\pm$ 0.4\\
        -random-medium-expert & 19.6 $\pm$ 2.5 & 41.4 $\pm$ 1.1 & 45.5 $\pm$ 1.3 & 46.9 $\pm$ 0.5\\
    \bottomrule
    \end{tabular}
    
\end{table}

\newpage
\section{Decomposition comparisons}
\label{sec: decomposition comparisons}
    In this Section we compare the DecQN decomposition to two alternative methods that can be used for factorisable discrete action spaces. The first is based on the Branching Dueling Q-Network (BDQ) proposed by \citet{tavakoli2018action}. Using our notation, each utility function is considered its own independent Q-function, i.e.
        \begin{equation}
            Q^i_{\theta_i} (s, a_i) = U_{\theta_{i}}^i(s, a_i) \; .
        \end{equation}
    Each Q-function is trained by bootstrapping from its own target, and no decomposition is used. That is, the target for $Q_{\theta_i}^i(s, a_i)$ is given by $y = r + \gamma \max_{a'_i \in \mathcal{A}_i} Q^i_{\bar{\theta}_i}(s', a'_i)$. The findings of \cite{ireland2023revalued} demonstrate that, in the online setting, BDQ is unable to match the performance of DecQN. This is likely caused by the fact that, as each sub-action space is now learnt independently, the effects of other sub-actions are treated as effects of the environment dynamics. Due to the fact that each agent is continually updating its own policy, this leads to non-stationary environment dynamics, making the learning problem much more challenging. 

    We also consider an alternative value-decomposition technique to the mean, namely the sum. That is, we replace the mean operator in Equation \ref{Q-Decomp} with the sum operator:
    \begin{equation}
        Q_\theta(s, \mathbf{a}) = \sum^N_{i=1} U^i_{\theta_i}(s, a_i) \; .
    \end{equation}
    Whilst this may seem a subtle change, \cite{ireland2023revalued} proved that the mean and variance of the learning target under this decomposition are both higher than DecQN. Empirical experiments by \citet{seyde2022solving, ireland2023revalued} also confirm the inferior performance of the sum decomposition compared to the mean. 

    In Table \ref{tab:DecQNvsBDQ} we can see that the sum decomposition is less performant in each of the tasks and datasets than the mean. For BDQ, we see that whilst in some cases performance is better than using the sum decomposition, it is generally still less performant than using the mean decomposition.  Owing to these results, we focus on the mean decomposition in our main work. 

    \begin{table}[ht]
        \centering
        \caption{Individual performance comparison for DecQN-CQL using mean and sum decompositions and BDQ-CQL for $n=3$. Figures are mean normalised scores $\pm$ one standard error, with 0 and 100 representing random and expert policies, respectively. \label{tab:DecQNvsBDQ}}
            \begin{tabular}{lccc}
                \toprule
                    Environment  -dataset & DecQN-CQL (Mean) & DecQN-CQL (Sum)       & BDQ-CQL         \\
                    \midrule
                    HumanoidStand         &                   &                      &                 \\
                    -expert               & 109 $\pm$ 1.9     &  95.1 $\pm$ 1.8      & 105.5 $\pm$ 0.7 \\
                    -medium-expert        & 104.7 $\pm$ 1.9   &  86.1 $\pm$ 2.8      & 94.0 $\pm$ 5.8  \\
                    -medium               & 51.4 $\pm$ 0.3    &  44.6 $\pm$ 0.8      & 48.2 $\pm$ 0.5  \\
                    -random-medium-expert & 42.7 $\pm$ 0.9    &  36.3 $\pm$ 1.6      & 43.2 $\pm$ 0.5  \\
                    \midrule
                    DogTrot               &                   &                      &                 \\
                    -expert               & 99.5 $\pm$ 0.7    &  93.1 $\pm$ 1.3      & 94.9 $\pm$ 1.2  \\
                    -medium-expert        & 84.8 $\pm$ 3.7    &  81.4 $\pm$ 2.8      & 75.7 $\pm$ 3.6  \\
                    -medium               & 46.5 $\pm$ 0.5    &  44.3 $\pm$ 0.3      & 48.5 $\pm$ 0.5  \\
                    -random-medium-expert & 43.4 $\pm$ 0.5    &  39.9 $\pm$ 0.9      & 37.9 $\pm$ 2.3  \\
                \toprule
            \end{tabular}
    \end{table}

\end{document}